\documentclass{article}
\usepackage{geometry}
\usepackage{authblk}
\usepackage{natbib}
\usepackage[utf8]{inputenc} 
\usepackage[T1]{fontenc}    
\usepackage{hyperref}       
\usepackage{url}            
\usepackage{booktabs}       
\usepackage{amsfonts}  
\usepackage{amsmath}
\usepackage{amsthm}
\usepackage{adjustbox}

\newtheorem{definition}{Definition}

\newtheorem{corollary}{Corollary}

\usepackage{nicefrac}       
\usepackage{microtype}      
\usepackage{xcolor}         
\usepackage{graphicx}       
\usepackage{subcaption}     
\usepackage{bbm}
\usepackage{paralist}
\usepackage{microtype}
\usepackage{bbm}
\usepackage{amssymb}
\usepackage{pdfpages}
\usepackage{fancyhdr}
\usepackage{graphicx}
\usepackage{caption}
\usepackage{multirow}
\usepackage{stmaryrd}
\usepackage[parfill]{parskip}
\usepackage{mathtools}
\usepackage{cases}
\usepackage{comment}
\usepackage{paralist}
\usepackage{breqn}
\usepackage{color}
\usepackage{algorithm, algorithmic}
\usepackage{appendix}
\usepackage{xspace}
\usepackage{enumitem}
\usepackage{gensymb}
\usepackage[english]{babel}
\usepackage{xcolor}
\usepackage{todonotes}
\usepackage{wrapfig}
\usepackage{letltxmacro}
\usepackage{breqn}
\usepackage{caption}

\let\svthefootnote\thefootnote
\newcommand\blfootnote[1]{%
  \let\thefootnote\relax%
  \footnotetext{#1}%
  \let\thefootnote\svthefootnote%
}

\newcommand{\sltd}{SLTD\xspace}
\newcommand{\mozannar}{{Mozannar et. al.}\xspace}
\newcommand{\madras}{{Madras et. al.}\xspace}
\newcommand{\shaping}{{Augmented-MDP}\xspace}
\newcommand{\synthetic}{synthetic data\xspace}
\newcommand{\Synthetic}{Synthetic data\xspace}

\newcommand{\Disc}{Synthetic \xspace}
\newcommand{\Diabetes}{Diabetes \xspace}

\LetLtxMacro\oldttfamily\ttfamily
\DeclareRobustCommand{\ttfamily}{\oldttfamily\csname ttsize\endcsname}
\newcommand{\setttsize}[1]{\def\ttsize{#1}}%
\setttsize{\footnotesize}%

\DeclareMathAlphabet\mathbfcal{OMS}{cmsy}{b}{n}

\newcommand{\defref}[1]{{Definition}~\ref{#1}}

\def\to{{\,\rightarrow\,}}

\mathchardef\mhyphen="2D

\providecommand{\trans}[1]{{#1^\top}}

\providecommand{\dot}[2]{{\trans{#1}#2}}

\newcommand{\norm}[1]{{ \left\lVert\right\rVert }}

\newcommand{\vertiii}[1]{{\left\vert\kern-0.25ex\left\vert\kern-0.25ex\left\vert #1
    \right\vert\kern-0.25ex\right\vert\kern-0.25ex\right\vert}}

\newcommand{\vect}[1]{{\boldsymbol{#1}}}

\def\bmu{\vect{\mu}}

\def\btheta{\vect{\theta}}

\def\ba{{\mathbf{a}}}

\def\bs{{\mathbf{s}}}

\def\cA{\mathcal{A}}

\def\cD{\mathcal{D}}

\def\cM{\mathcal{M}}
\def\cN{\mathcal{N}}

\def\cP{\mathcal{P}}

\def\cS{\mathcal{S}}

\def\bcM{\mathbfcal{M}}

\title{Learning-to-defer for sequential medical decision-making under uncertainty}

\newcommand*\samethanks[1][\value{footnote}]{\footnotemark[#1]}
\author[1]{Shalmali Joshi\thanks{Equal Contribution}\thanks{Corresponding Author:shalmali@seas.harvard.edu}}
\author[1,2]{Sonali Parbhoo\samethanks[1]}
\author[1]{Finale Doshi-Velez}
\affil[1]{Harvard University (SEAS)}
\affil[2]{Imperial College London}

\begin{document}

\maketitle

\begin{abstract}
\noindent Learning-to-defer is a framework to automatically defer decision-making to a human expert when ML-based decisions are deemed unreliable. Existing learning-to-defer frameworks are not designed for sequential settings. That is, they defer at every instance independently, based on immediate predictions, while ignoring the potential long-term impact of these interventions. As a result, existing frameworks are myopic. Further, they do not defer adaptively, which is crucial when human interventions are costly. In this work, we propose Sequential Learning-to-Defer (SLTD), a framework for learning-to-defer to a domain expert in sequential decision-making settings.  Contrary to existing literature, we pose the problem of learning-to-defer as model-based reinforcement learning (RL) to i) account for long-term consequences of ML-based actions using RL and ii) adaptively defer based on the dynamics (model-based). Our proposed framework determines whether to defer (at each time step) by quantifying whether a deferral now will improve the value compared to delaying deferral to the next time step. To quantify the improvement, we account for potential future deferrals. As a result, we learn a pre-emptive deferral policy (i.e. a policy that defers early if using the ML-based policy could worsen long-term outcomes). Our deferral policy is adaptive to the non-stationarity in the dynamics. We demonstrate that adaptive deferral via \sltd provides an improved trade-off between long-term outcomes and deferral frequency  on synthetic, semi-synthetic, and real-world data with non-stationary dynamics. Finally, we interpret the deferral decision by decomposing the propagated (long-term) uncertainty around the outcome, to justify the deferral decision.
\end{abstract}

\section{Introduction}
\label{sec:intro}

Machine learning (ML) has the potential to be deployed for decision-making in complex domains such as healthcare, lending, and legal systems. In many cases, ML-based policy may not generalize to situations not encountered during training. In practice, it may be safer to defer to a human expert when using the ML policy may not improve outcomes or cause active harm. Automatically deferring to a human expert is called `Learning-to-defer.' Earlier works have considered the problem of learning-to-defer in non-sequential settings \citep{mozannar2020consistent,madras2017predict}..

In situations such as managing health, however, two key challenges remain. First, deferral decisions can significantly alter long-term outcomes. Thus modeling the long-term outcome is critical to decide \emph{when} to defer to an expert. Deferring too late may lead to unintended and irreversible harm. Deferring too early may increase the burden on the human expert. Second, when human interventions (after deferral) are costly, learning-to-defer \emph{adaptively} and only when critical is crucial. To defer adaptively, we need a well-characterized model of the environment, a challenging estimation issue, especially under non-stationarity, i.e., when the dynamics of the environment change over time.  

Existing learning-to-defer methods defer based on immediate outcomes e.g.~\citet{mozannar2020consistent,madras2017predict, Gennatas4571}, and are therefore myopic. Further, the objective to defer is to improve the performance of some prediction tasks (such as the ability to predict a patient outcome). These frameworks either defer based on the probability of correct short-term prediction or characterizing the trade-off of paying a cost (to defer).  Instead, interventions based on an ML system can have long-term consequences that are crucial to the model. 
Further, in many cases, merely deferring to optimize for decision/ prediction accuracy in a supervised learning setting does not suffice to improve long-term outcomes. Existing approaches also do not leverage the potential of modeling the environment to defer \emph{adaptively}, especially beneficial if the environment is non-stationary.  

 \begin{figure}[!htbp]
         \centering
         \includegraphics[width=0.5\textwidth]{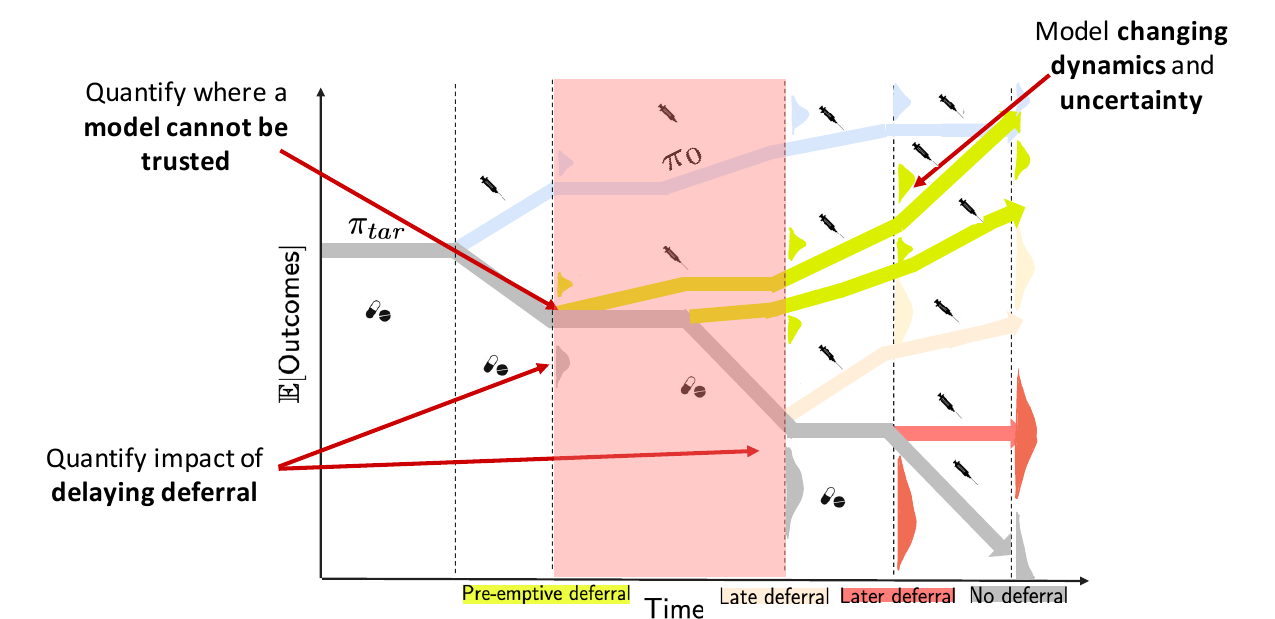}
         \caption{Overview of deferral strategies in a medical setting. The target policy $\pi_{\text{tar}}$ recommends continuing a pill-based therapy, while the domain expert ($\pi_{0}$) suggests switching to a shot. SLTD (green) defers in the shaded region where $\pi_{\text{tar}}$ is unlikely to improve expected future outcomes. To do so, SLTD models changing dynamics, and accounts for the impact of delayed deferral to identify and defer in the shaded region.} 
     \label{fig:overview}
\end{figure}

{\bf{Algorithmic Motivation.}} To address these challenges, we deviate significantly from existing learning-to-defer methods, which use a supervised learning framework. Instead, we model the learning-to-defer problem for sequential settings as offline model-based reinforcement learning (RL). \sltd is the first RL-based learning-to-defer framework. We focus on settings where online experimentation is prohibitive for safety reasons, such as healthcare. 

We assume access to batch data collected by human experts (such as clinicians) using a behavior policy. Our goal is to learn a deferral policy with respect to a fixed ML-based policy (called the target policy). \sltd decides whether or not to defer (to the expert behavior policy) at each instance by modeling the impact of \emph{delaying deferral} (by one time step) on the long-term outcomes. \sltd defers if delaying deferral does not improve outcomes compared to deferring in the current instance. To quantify long-term outcomes, we also account for all future deferrals. In doing so, \sltd precisely identifies the regions of the state space where the target ML-based policy will not improve outcomes. As a result, our method is pre-emptive,   i.e.,  \emph{it defers in all regions where the ML policy is unlikely to improve long-term outcomes}. See Figure \ref{fig:overview} for a conceptual overview of \sltd.  

Human expert interventions are often costly. Hence deferring too often is not desirable. To defer \emph{adaptively}, we propose to leverage an estimate of the environment dynamics and the associated uncertainty. Modeling the dynamics allows us to reliably quantify the impact of delaying deferral on long-term outcomes, which is particularly beneficial in non-stationary settings. 
We show that modeling the non-stationarity provides a better trade-off of improving outcomes versus the frequency of deferrals. In a myopic environment (i.e., when the effect of interventions are observed in the near future), it may seem unnecessary to model the dynamics. However, we demonstrate that deferral methods that defer myopically, based on immediate outcomes still benefit from modeling the dynamics, and consequently the impact of potential future myopic deferrals.   
When \sltd defers, human experts can benefit from an additional justification of the deferral decision to determine potential interventions. Hence, we also interpret \sltd's decision to defer at any given time by quantifying the long-term uncertainty in the outcome and decomposing the sources of uncertainty. We justify how the decomposition can guide experts to potential interventions.  
 
 {\bf{Clinical Motivation.}} We are motivated by clinical settings where a target policy is learned from
batch data to work well across multiple institutions. Such a policy may perform well on average, but when
deployed to a new environment,  encounter a different or an evolving patient population. Clinicians
might also follow a slightly different treatment protocol than this learned policy. Regulatory constraints
may prevent significantly adapting our target policy completely to the new site. In this case, it is safer to
leverage batch data from the new site to quantify when it is reliable to deploy the learned target policy. In
situations where the policy does not improve outcomes compared to the human expert, exacerbated by challenges
like non-stationarity, it is safer to defer to human experts. Beyond healthcare, our work is applicable in many
safety-focused, data-scarce, non-stationary settings where online policy improvement is not allowed due to
ethical or practical constraints. 

 \section{Related Work \& Background}\label{sec:relatedwork}

\paragraph{Mixture-of-Experts (MoE).}
Many methods focus on deciding to deploy two or more policies. For example, \citet{jacobs1991adaptive,jordan1994hierarchical} switch between different policies in decision-making by partitioning the input space into regions assigned to different specialized sub-models. Variants of this framework enforce an explicit preference for a specific expert, e.g.,  a human expert, and train other experts to complement the human expert \citep{pradier2021preferential}. In sequential settings, \citet{parbhoo2017combining, gottesman2019combining,parbhoo2018improving} combine parametric and non-parametric experts to learn more accurate estimates of the value function. On the other hand, we focus on \emph{deferral to human experts when future outcomes using the current ML-based policy are potentially undesirable}. Further, we defer based on explicitly quantifying the impact of delayed deferral to decide \emph{when} to defer. 

\paragraph{Policy Improvement with Expert Supervision.}
\citet{NEURIPS2020_daf64245} use hypothesis testing to assess whether, at each state, a policy from a human expert would improve value estimates over a target policy \emph{during training} to improve the target policy. In contrast, our work identifies the value of \emph{delaying deferral to a human expert at test time}. Improvements using expert supervision are unlikely to be always feasible due to safety and regulatory constraints. Learning-to-defer with respect to a fixed target policy is crucial as a safeguard. Some works focus on safe policy improvement in a non-stationary MDP setting \citep{ chandak2020optimizing,chandak2020towards}.  \citet{chandak2020towards} assume that the non-stationarity is governed by an exogenous process, and so past actions do not impact the underlying non-stationarity. Our work differs in two ways: first, we argue that model misspecification, specifically ignoring non-stationarity induced by (deferral) actions, affects the likelihood of future deferrals. 
Accounting for this non-stationarity is crucial to avoid costly deferrals.  Second, we incorporate human expertise by explicitly measuring the impact of delaying deferral. 

\paragraph{Learning-to-defer to Human Expertise.} \citet{madras2017predict, mozannar2020consistent} propose supervised models to defer to the expert. 
Here, the classifiers are trained on the samples of an expert's decisions. \citet{madras2017predict} train a separate rejection and prediction function, while \citet{mozannar2020consistent} learn a joint predictor for all targets and deferral. \citet{madras2017predict} is conceptually closer to our work but in a non-sequential setting. Other approaches such as \citet{raghu2019algorithmic, wilder2020learning} first train a standard classifier on the data and then compute uncertainty estimates for this classifier and the human expert.  
The models defer to the expert if the model is highly uncertain or can significantly benefit from deferral. \citet{liu2021incorporating}  
incorporate uncertainty in Learning-to-Defer algorithms for classification tasks.
Instead, we focus on learning-to-defer in non-stationary, sequential settings.

\paragraph{Decomposing Uncertainty for Interpreting Policies.} Uncertainty, if well calibrated can help decision-makers understand the failure modes of a model~\citep{bhatt2020uncertainty,tomsett2020rapid,zhang2020effect}. Several methods estimate predictive uncertainty in  ML~\citep{pmlr-v48-gal16,guo2017calibration}. Here, we focus on capturing the \emph{propagated uncertainty} in sequential settings to interpret deferral decisions. We interpret the (different) sources of propagated uncertainty when \sltd defers to the expert. Decomposing the sources of uncertainty into modeling and irreducible uncertainty over predictions has been explored in classification and prediction settings~\citep{yao2019quality,depeweg2018decomposition} but remains significantly under-explored for sequential settings.  

\paragraph{Background and Notation.} We consider our environment to be a finite horizon MDP defined by $\cM \equiv (\cS, \cA, \cP, r, p_0)$ where $\cS$ indicates the state-space, $\cA$ indicates the action-space, $\cP$ the transition dynamics, $r: s \times a \to \mathbb{R}_{+}$ the reward function, $p_0$ the initial state distribution. The action-space is assumed to be discrete, while the state space can be discrete or continuous.  Any intervention policy (usually stochastic in our case) is given by $\pi=: \cS \times \cA \to [0,1]$. We consider a non-stationary environment such that the dynamics at any time $t$ are  governed by a specific MDP $\cM_{t}$. Thus the environment is a sequence of MDPs. 
We assume the existence of a true set of non-stationary dynamics governing all episodes and denote it by $\mathbfcal{M}^* := \{\cM_{t}^*\}_t$. In the rest of the draft, $\mathbfcal{M} := \{\cM_{t}\}_t$ denotes an estimate of the true dynamics $\bcM^*$. Let $T$ be the episode-length.  The value of a policy $\pi$ at $t$ is given by $V_{\pi,t}^{\bcM}(s) = \mathbb{E}_{{\bcM}}[\sum_{j=t}^{T} r^{j} (s, a) | s_{t}=s, \pi]$.  
The action value is given by $Q^{\bcM}_{\pi, t}(s, a) = r(s, a) + \sum_{s' \in \cS} \cP(s'|s,a)V_{\pi, t}^{\bcM}(s')$. 

\section{Sequential Learning-to-Defer}\label{sec:method}


{\bf{Problem Setup.}} Assume we are given a policy $\pi_{\text{tar}}$ that may be learned from batch data from one or more environments. 
 $\pi_{\text{tar}}$ is intended to be deployed in a new environment. 
 %
 We have access to batch data, denoted by $\cD^* = \{s_{i,0}, a_{i,0}, r_{i,0}, \cdots, s_{i, T}, a_{i, T}, r_{i, T}\}_{i=1}^N$ collected in the new non-stationary environment $\bcM^* = \{\bcM_t^*\}_t$, from some (potentially non-stationary) behavior policy $\pi_0$. Here $N$ denotes the number of episodes. Our goal is to learn a deferral policy $g_{\pi_{\text{tar}}}(s, t): \cS \times T \to \{0, 1\}$ (where $1$ corresponds to defer or $\perp$) with respect to $\pi_{\text{tar}}$ to defer to the expert policy $\pi_0$. 
 
 Deferral to the expert is denoted by the action $\perp$. That is, we will augment the action space of existing MDP $\bcM^*$ to include a new deferral action $\cA_{\perp}:= \cA\, \cup \perp$. At every step, the agent decides whether or not to defer. If the agent defers, $\pi_0$ will be deployed for that time step. 
 We describe the formulation assuming strict adherence to $\pi_0$ at deferral to emphasize other aspects of our contribution such as the impact of non-stationarity and how to account for relevant sources of uncertainty to compare outcomes. \sltd can easily account for the uncertainty of expert actions in the framework.

In practice, the target policy $\pi_0$ may not uniformly improve over $\pi_{\text{tar}}$ for all states. That is guaranteeing  that $V_{\pi_{\text{tar}},0}^{\bcM^*}(s) \geq V_{\pi_{0},0}^{\bcM^*}(s)$ for all $s \in \cS$,  is challenging. Even when fine-tuning is allowed, it is challenging to ensure that the target policy is indeed better than $\pi_0$ in all regions of the state space. Hence, we would like to get the best of both worlds. We can deploy $\pi_{\text{tar}}$, to reduce the costs of relying on human expertise, and learn to automatically defer to the costlier policy $\pi_{0}$  (i.e. human expert) when relying on $\pi_{\text{tar}}$ does not improve outcomes. In regions of the state-space where the value of $\pi_{\text{tar}}$ is lower than $\pi_{0}$, it is better to defer to the human as a ``safety protocol''.


\paragraph{\sltd.} 
To determine whether to defer at each time step, we quantify whether deferring (relying on the behavior policy) or not deferring (using the target policy) at the current time step improves the long-term outcome. Long-term outcomes are affected by potential future deferrals. Thus comparing the consequences of deferring versus relying on the ML policy at the current instance is equivalent to comparing the impact of deferring now versus delaying deferral by one time step. 

Future deferrals imply that some unknown mixture of $\pi_{\text{tar}}$ and $\pi_0$ is used in the future. We denote such a mixture policy as $\pi_{\text{mix}}$. To minimize cumbersome notation, we denote a policy where $\pi_{\text{tar}}$ is deployed at instance $t$ and $\pi_{\text{mix}}$ in the future as: $\pi_{\text{tar}(t),\text{mix}(t_+)}$. Similarly, if we defer \emph{now}, then the policy that is deployed at time $t$ is $\pi_0$ , and $\pi_{\text{mix}}$ in the future. We denote this mixture as $\pi_{0(t),\text{mix}(t_+)}$. Thus, at any instance $t$, we want to defer if $V_{\pi_{\text{tar}(t), \text{mix}(t_+)}}^{\bcM}(s) < V_{\pi_{0(t), \text{mix}(t_+)}}^{\bcM}(s)$. Note that we consider deferral to $\pi_0$ as a costly one. 
This is accounted through a constant cost $c>0$ in terms of the value. That is, deferral incurs cost $c$ and the resulting value is: $V_{\pi_{0(t), \text{mix}(t_+)}}^{\bcM}(s)-c$. We can now formalize our stochastic deferral policy:
 
\begin{definition}\label{def:deferral}
Let $\pi_{\text{tar},t}$ be such that such that there exists $\texttt{s}^t \subseteq \cS \, \forall t \in \{0, 1, \cdots, T\}$ where $P(V_{\pi_{\text{tar}(t), mix(t_+)}}^{\bcM}(s) < V_{\pi_{0(t), mix(t_+)}}^{\bcM}(s)-c) > \tau$ for constant cost of deferral $c>0$ and threshold $\tau >0$, $\forall s \in \texttt{s}^t$. Then the deferral policy $g_{\pi_{\text{tar}}}(s,t) \triangleq  \mathbf{1}[P(V_{\pi_{\text{tar}(t), mix(t_+)}}^{\bcM}(s) < V_{\pi_{0(t), mix(t_+)}}^{\bcM}(s)-c)> \tau] \triangleq \mathbf{1}[\tilde{g}_{\pi_{\text{tar}}}(s,t) > \tau]$. 
\end{definition}

\begin{corollary}\label{corollary:preemptive}
By Definition \ref{def:deferral}, $g_{\pi_{\text{tar}}}(s,t)$, includes the earliest time in the episode where $\tilde{g}_{\pi_{\text{tar}}}(s,t) \triangleq P(V_{\pi_{\text{tar}(t), \text{mix}(t_+)}}^{\bcM} < V_{\pi_{0(t), \text{mix}(t_+)}}^{\bcM}-c)> \tau$. Thus, $g_{\pi_{\text{tar}}}(s,t)$ is a pre-emptive deferral policy.
\end{corollary}

The cost $c$ determines how conservative \sltd is and trades-off frequency of deferral to the value attained.
 This parameter should be tuned by domain experts aware of the trade-off and risks involved. For instance, in a critical care setting, we may be more conservative and use a smaller $c$ than in a chronic care situation. $\tau$ is a safety threshold on the probability  of worse outcome beyond which we deem that deferral is necessary. 

\defref{def:deferral} indicates that to reliably learn the deferral policy, we need to estimate $\tilde{g}_{\pi_{\text{tar}}}(s,t) \triangleq  P(V_{\pi_{\text{tar}(t), mix(t_+)}}^{\bcM}(s) < V_{\pi_{0(t), mix(t_+)}}^{\bcM}(s)-c)$.  
To estimate this probability, we should model all sources of uncertainty in the system, including the non-stationary dynamics, and the uncertainty associated with our modeling assumptions. We use a Bayesian RL approach to account for all sources of uncertainty. We motivate this by first describing our dynamic programming approach to learn-to-defer.  

Our dynamic programming procedure maintains an estimate of the deferral probability  $\tilde{g}_{\pi_{\text{tar}}}(s,t)$ and refines it as we train on the batch data. Given an estimate of $\tilde{g}_{\pi_{\text{tar}}}(s,t)$, we outline the procedure to i) estimate the value under mixture policies corresponding to deferral (and delayed deferral), ii) modeling the probability of improvement under various sources of uncertainty, and finally iii) obtaining a new estimate of the deferral probability $\tilde{g}_{\pi_{\text{tar}}}(s,t) \forall s \in \cS$ at the given time $t$ using i) and ii). We then bootstrap this procedure over our batch data  to refine our deferral probabilities. We describe the procedure for the discrete setting. 

\begin{algorithm}[!htbp]
\caption{Sequential Learning to Defer}
\label{alg:rl_ltd}
\begin{algorithmic}
\STATE{\bfseries Input:} $\cD^*$, behavior policy $\pi_0$, target policy $\pi_{\text{tar}}$.
\STATE{Estimate Posterior Distributions $\{\cM_t \triangleq p_t(\cdot| \cD^*)\}_{t=0}^{T}$ (posteriors over rewards not shown here)}
\STATE {\bfseries Initialization:} Deferral function $g_{\pi_{\text{tar}}}(s,t) =0 \, $ for all $s \in \cS$ and $t \in \{1, 2, \cdots, T\}$.
\FOR{$n \in \text{BOOTSTRAPS}(\cD^*)$}
\STATE{Sample ${\bcM_k} \triangleq \{\cM_{k,t} \sim p_{t}(\dot | \cD*) \} \forall t \in \{1, 2, \cdots, T\}, \forall k \in \{1, 2, \cdots, K\}$}
 \FOR{$t \in \{T, T-1, \cdots, 1\}$}
 \FOR{$s \in \cS$}
  \STATE{Compute $V_{\pi_{\text{tar}(t), \text{mix}(t_+)}}^{\bcM}$,  $V_{\pi_{0(t), \text{mix}(t_+)}}^{\bcM}- c \, \forall \bcM$ }
  \STATE{{\small{$\tilde{g}_{\pi_{\text{tar}}}(s,t) \leftarrow 
  \approx  
  \frac{1}{K}\sum_{\bcM_k \sim \{p_{t'}(\cdot \vert \cD)\}_{t'=t}^T}[\mathbf{1}(V_{\pi_{\text{tar}(t), \text{mix}(t_+)}}^{\bcM_k} < V_{\pi_{0(t), \text{mix}(t_+)}}^{\bcM_k}-c)]$}}}
  \ENDFOR
  \ENDFOR
  \ENDFOR
  \STATE {\bfseries return} $g_{\pi_{\text{tar}}}(s,t) = \mathbf{1}(\tilde{g}_{\pi_{\text{tar}}}(s,t) > \tau) \forall s,t \in \cS \times \{1,2,\cdots, T\}$
  \end{algorithmic}
  \end{algorithm}

\paragraph{Estimating Value function.} At any instance we defer based on current estimates of $g_{\pi_{\text{tar}}}(s,t)$ (or equivalently $\tilde{g}_{\pi_{\text{tar}}}(s,t)$). We sample actions from $\pi_{\text{tar}}$ if  $g_{\pi_{\text{tar}}}(s,t) = 0$ and $\pi_0$ otherwise (equivalent to $\perp$). Note that the current estimate of  $g_{\pi_{\text{tar}}}(s,t)$ determines the future mixture policy as well. We now estimate the value of the mixture policies using the Bellman Equation of the state and action value functions. For the mixture policy $\pi_m \triangleq \pi_{tar(t), mix(t_+)}$ (corresponding to no deferral at $t$), the Q-function is:
{\small{
\begin{equation}\label{eq:q_bellman}
    Q^{\bcM}_{\pi_m, t}(s, a) = r(s, a) + \sum_{s' \in \cS} \cP^{\bcM}(s'|s,a)V_{\pi_{\text{mix}(t_+)}, t+1}^{\bcM}(s')
\end{equation}}}
and the Value function is:
{\small{
\begin{equation}\label{eq:v_bellman}
    V_{\pi_m,t}^{\bcM}(s) = \sum_{a \in \cA} \pi_{\text{tar}(t)}(a\vert s) Q^{\bcM}_{\pi_m, t}(s,a)
\end{equation}}}
Similarly for the mixture policy if we defer at $t$. 

\paragraph{Estimating the probability of improving outcomes by delaying deferral.}
 At each instance $t$, for all states $s$, we can estimate the indicator function $\mathbf{1}[V_{\pi_{\text{tar}(t), \text{mix}(t_+)}}^{\bcM^*}(s) < V_{\pi_{0(t), \text{mix}(t_+)}}^{\bcM^*}(s)-c]$ given an estimate of $\tilde{g}_{\pi_{\text{tar}}}$ as described above. However, we do not have access to the true dynamics $\bcM^*$. In batch settings, such as ours, we often estimate the dynamics using maximum-likelihood estimation. Such methods make specific assumptions about the distribution governing the dynamics. 
 Our assumptions about the dynamics may be incorrect resulting in potential misspecification of our dynamics model. This increases the uncertainty in the outcome and potentially over-estimates the probability that relying on the model may improve outcomes. To account for this additional source of uncertainty, we use a Bayesian RL approach. We describe the procedure for the dynamics. The procedure for rewards follows an analogous process.  
 
  Suppose the parameters of the distributions governing the dynamics are denoted by $\theta_t \, \forall t \in \{0, \cdots, T\}$. We denote the full set of parameters by $\btheta = \{\theta_t\}_{t}$. We assume a prior distribution over the parameters of the distribution governing the dynamics $\cP^{\bcM}_{\theta}(s'|s, a)$ and the rewards $r(s, a)$. Given batch samples $\cD^*$, we can estimate the posterior distribution over the non-stationary MDPs and rewards using Bayesian inference: $$p(\btheta \vert \cD^*) \propto p(\cD^* \vert \btheta)p(\btheta)$$ 
 
 More specifically, we assume conjugate priors for our parameters $\btheta$. By relying on conjugate priors in our inference, the parameters of posterior distributions over the dynamics and rewards are obtained in closed form. For discrete state dynamics (and rewards), we assume a Dirichlet prior distribution and model the observations $p(\cD^* \vert \btheta)$ using a Multinomial distribution. For continuous states, $p(\cD^* \vert \btheta)$ is assumed to be normally distributed with $\btheta$ being the mean and variance parameters. The prior distributions over the mean and precision (inverse of the variance) is the Normal-gamma prior. This is a domain-dependent choice and \sltd is agnostic so long as we can sample from the posterior distributions of the learned model dynamics.  A  detailed derivation of how the data is leveraged to estimate the posterior distributions over the dynamics are provided in Appendix~\ref{app:posterior}. By allowing flexibility of modeling the dynamics via Bayesian RL, we can account for uncertainty over our modeling assumptions.  
 
 Finally, based on our assumption that the non-stationary environment is governed by a sequence of MDPs, we estimate the MDP for each time step independently from batch data. This allows us to make fewer assumptions about the \emph{type} of non-stationarity. Any additional domain knowledge about the nature of non-stationarity can be leveraged for data efficiency. We can now estimate the impact of delayed deferral by sampling non-stationary MDPs from our posterior distributions and averaging to obtain our final probability:
 {\small{
 \begin{equation}\label{eq:G_est}
 \begin{aligned}
   &\tilde{g}_{\pi_{\text{tar}}}(s,t) \triangleq P(V_{\pi_{\text{tar}(t), \text{mix}(t_+)}}^{\bcM^*}(s) < V_{\pi_{0(t), \text{mix}(t_+)}}^{\bcM^*}(s)-c) \\
   &= \mathbb{E}_{\bcM \sim p(\cdot |\cD^*)} [\mathbf{1}[V_{\pi_{\text{tar}(t), \text{mix}(t_+)}}^{\bcM}(s) < V_{\pi_{0(t), \text{mix}(t_+)}}^{\bcM}(s)-c]] \\
   &\approx \frac{1}{K}\sum_{\bcM_k \sim \{p_{t'}(\cdot \vert \cD^*)\}_{t'=t}^T} \mathbf{1}[V_{\pi_{\text{tar}(t), \text{mix}(t_+)}}^{\bcM_k}(s) < V_{\pi_{0(t), \text{mix}(t_+)}}^{\bcM_k}(s)-c]\\
 \end{aligned}
 \end{equation}}}
 where the second line comes from the definition due to the randomness over the dynamics, and the last term comes from approximating the expectation using $K$ samples from the posterior distribution of the dynamics $p(\cdot \vert \cD^*)$. Thus, for every instant $t$, in a given state $s$, our deferral policy $g_{\pi_{\text{tar}}}(s,t)$ is given by, $g_{\pi_{\text{tar}}}(s,t) := \mathbf{1}[\tilde{g}_{\pi_{\text{tar}}}(s,t)> \tau]$.  

\paragraph{Dynamic Programming to estimate $g_{\pi_{\text{tar}}}(s,t)$.} 
Our dynamic programming procedure is summarized in Algorithm~\ref{alg:rl_ltd}. We initialize $g_{\pi_{\text{tar}}}(s,t) = 0$ for all $s \in \cS$. We estimate $V_{\pi_{\text{tar}(t), \text{mix}(t_+)}}^{\bcM}(s) ,  V_{\pi_{0(t), \text{mix}(t_+)}}^{\bcM}(s)$ for a given $t$ using Bellman Equations \ref{eq:q_bellman} and \ref{eq:v_bellman}. Following that, we can update our estimate of $g_{\pi_{\text{tar}
}}(s,t)$ using our posterior MDPs, i.e.  Equation~\ref{eq:G_est}. We repeat (over $t$) using the \emph{updated} estimates of $g_{\pi_{\text{tar}}}(s,t)$.


\section{Decomposing the uncertainty at deferral}\label{sec:unc_decomp}
\sltd defers at time $t$ because the probability that relying on $\pi_{\text{tar}}$   improves the outcome is below our safety threshold, i.e. \sltd is uncertain of an improved outcome. Conveying this uncertainty can help the domain expert take over decision-making. We interpret this deferral decision in terms of the total and decomposed uncertainty on long-term outcomes. 
We convey two different sources of uncertainty at deferral. First, we consider \emph{epistemic/modeling uncertainty}, which captures whether our model specification has resulted in high uncertainty and the \emph{aleatoric uncertainty} which mainly results from the stochasticity of the environment itself. A high relative value of the former suggests that adding more data to train \sltd can improve the confidence of the model. High aleatoric uncertainty suggests that the environment itself is highly variable leading to the lack of confidence in relying on $\pi_{\text{tar}}$. 

Concretely, let $t_{d}$ be a time when \sltd defers. The agent is in state $s_{t_d}$. We are interested in the reward (and uncertainty over the reward) at time $T$  due to deferral at $t_{d}$, i.e., $\mathbb{E}[r_{T}|s_{t_{d}}, \mu_{t_d}, \pi_{0(t_d), mix(t_d+)}]$. 
 We denote the posterior MDP samples for any state-action pair by $\mu_{t}$. 
 The variability in these samples captures modeling uncertainty. The dynamics parameters are denoted by $\theta_t (s, a)$ for each state-action pair. First, we sample the parameters of the dynamics from posterior distribution $p(\theta_{t'} | \cD^*)$, followed by sampling the MDPs $\mu_{t'} \sim p(\mu_{t'} | \theta_t'(s_{t'}, a_{t'}))$. 
  Once we defer, we sample actions from $\pi_0$ at time $t'=t_d$ and $\pi_{\text{mix}}$ for $t'> t_d$ where the mixture probability is determined by the learned $g_{\pi_{\text{tar}}}$ for future deferrals. The expected long-term outcome is given by:
{\small{
\begin{align*}
    \begin{split}
        &\mathbb{E}[r_{T}|s_{t_{d}}, \mu_{t_d}] = 
        \int_{s_{t_{d}+1}}^{s_{T}} \int_{a_{t_{d}}}^{a_{T}} \int_{\mu_{t_{d}+1}}^{\mu_{T}} \int_{\theta_{t_d}}^T r(s_T, a_T) \times 
        \prod_{t'={t_{d}+1}}^T p_{t'}(s_{t'} | \mu_{t'}) p_{t'}(\mu_{t'} | \theta_t'(s_{t'}, a_{t'})) 
        \pi_{t'}(a_{t'}| s_{t'}) p_{t'}({\theta_{t'}}| \cD) d\bs d\ba  d\bmu d{\btheta} 
    \end{split}
\end{align*}
}}

Integrands are written in short-hand: $\bs = \{s_{t_d+1}, s_{t_d+2}, \cdots, s_{T}\}$ (analogously for other quantities). We maintain one estimate of parameter $\theta_{t'}$ and sample $K$ MDPs $\mu_{t'}$ from this distribution. 
Thus, the epistemic uncertainty we capture is due to the uncertainty over dynamics under fixed parameters. 
The total uncertainty can now be decomposed using the law of total variance:
{\small{
\begin{align*}\label{eq:unc_decom}
    \begin{split}
     \underbrace{\text{Var}(r_{T} | s_{t_d}, \cD)}_{\text{Total Uncertainty}} = 
     \underbrace{\mathbb{E}_{\mu_{t_d} \sim p(\mu_{t_d}| \cD)}\big[\text{Var} (r_T| \mu_{t_d}, s_{t_d}, \cD)\big]}_{\text{Irreducible/ Aleatoric Uncertainty}}  + 
 \underbrace{\text{Var}_{\mu_{t_d} \sim p(\mu_{t_d}| \cD)} \big( \mathbb{E}[r_{T}| \mu_{t_d}, s_{t_d}, \cD]\big)}_{\text{Epistemic/Modeling  Uncertainty}}
    \end{split}
\end{align*}}}
The second term is the variance \emph{conditioned} on knowledge of the model $\mu_{t_d}$. 
 This is the \emph{propagated uncertainty due to modeling uncertainty at $t_d$} and can be reduced by data collection. The first term averages over the variance due to $\mu_{t_d}$ and captures \emph{propagated uncertainty due to aleatoric uncertainty at $t_d$}, which conveys stochasticity of the environment itself. This uncertainty can only be reduced by careful interventions at $t_d$. We estimate these using Monte-Carlo sampling. Additional details on the derivation are provided in Appendix~\ref{app:uncertainty_decomp}. As suggested before, a high \emph{propagated epistemic uncertainty} conveys that the current uncertainty of model prediction (of the dynamics) is high but could be improved if additional data could be collected. High \emph{propagated aleatoric uncertainty} indicates high variability in the dynamics that can only be reduced with careful interventions and is otherwise not manageable. 

\section{Experiments}\label{sec:exp}
\begin{figure*}[t]
   \centering
       \includegraphics[scale=0.145]{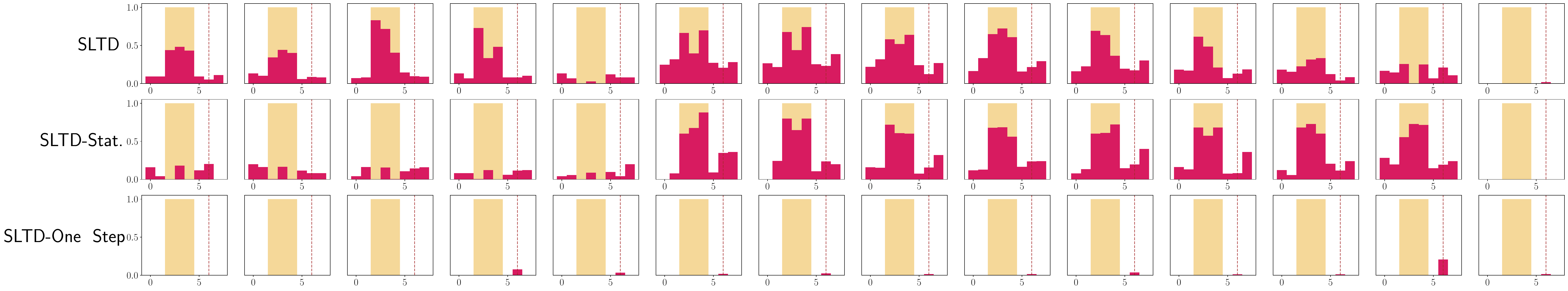}
    \caption{{\small{Learned deferral probabilities for \sltd (top row), \sltd stationary (second row),   \sltd one-step (third row), and \shaping (red dotted line) for $c=0$. Each row shows the deferral histogram $\tilde{g}_{\pi_{\text{tar}}}(s,t)$ over the length of the episode $T$ (major x-axis) for \Synthetic. Each box is the histogram with respect to the states at a fixed time. According to \Synthetic design, target policy always takes suboptimal actions in the yellow region. Further the dynamics changes over time so that the optimal action flips, as well as the noise increases, when $5 \leq t \leq 12$ requiring deferral more often. Thus, shaded yellow regions is
    the region of pre-emptive deferral.  \sltd, which models non-stationarity defers adaptively and early in the shaded yellow region (top row), and increases the deferral  probability when dynamics change. \sltd-stationary does not learn calibrated probabilities in the yellow region over time and only defers when the \emph{average} dynamics of the environment require deferral.  \sltd-one-step and \shaping (dotted red line) only defer in state $6$ when the reward is negative, and are not pre-emptive.}}} 
    \label{fig:def_policy_plot}
    \end{figure*}
We evaluate \sltd's ability to defer adaptively in sequential settings with respect to 
 a known and fixed $\pi_{\text{tar}}$ to the  expert policy $\pi_0$. We test the utility of: i)  deferring based on long-term outcomes, 
  ii) adaptively deferring by quantifying the impact of delaying deferral, i.e., in regions where delayed deferral can worsen  outcomes, iii)  modeling the non-stationarity on deferral frequency, iv) quantifying multiple sources of uncertainty to estimate the probability of different outcomes under delayed deferral. We test our method on \synthetic, a non-stationary diabetes simulator\footnote{Jinyu Xie. Simglucose v0.2.1 (2018) [Online]. Available: \url{https://github.com/jxx123/simglucose}. Accessed on: 07-24-2021.} modified from~\citet{chandak2020optimizing},  
  and real-world HIV data. 
\begin{table*}[!htbp]
\centering
\begin{adjustbox}{max width=\textwidth}
\begin{tabular}{@{}l|cccccccc@{}}
\toprule
\multirow{2}{*}{Method} &
  \multicolumn{2}{c}{\Disc} &
  \multicolumn{2}{c}{\Diabetes} &
  \multicolumn{2}{c}{HIV - Case Study I} &
  \multicolumn{2}{c}{HIV - Case Study II} \\ \cmidrule(l){2-9} 
 &
  \begin{tabular}[c]{@{}c@{}}Value\\ (mean $\pm$ 2 s.e.)\end{tabular} &
  \begin{tabular}[c]{@{}c@{}}Defer \\ Frequency\end{tabular} &
  \begin{tabular}[c]{@{}c@{}}Value\\ (mean $\pm$ 2 s.e.)\end{tabular} &
  \begin{tabular}[c]{@{}c@{}}Defer \\ Frequency\end{tabular} &
  \begin{tabular}[c]{@{}c@{}}Value\\ (mean $\pm$ 2 s.e.)\end{tabular} &
  \begin{tabular}[c]{@{}c@{}}Defer\\ Frequency\end{tabular} &
  \begin{tabular}[c]{@{}c@{}}Value\\ (mean $\pm$ 2 s.e.)\end{tabular} &
  \begin{tabular}[c]{@{}c@{}}Defer\\ Frequency\end{tabular} \\ \midrule
SLTD &
  \textbf{8.029 $\pm$ 0.039} &
  0.509 &
  \textbf{36.931 $\pm$ 0.166} &
  0.396 &
  \textbf{14.792 $\pm$ 0.267} &
  0.342 &
  \textbf{8.754 $\pm$ 0.125} &
  0.461\\
SLTD-Stat. &
  5.588 $\pm$ 0.048 &
  1.000 &
  34.326 $\pm$ 0.172 &
  1.000 &
  11.020 $\pm$ 0.230 &
  0.629 &
 4.291 $\pm$ 0.218 &
  0.317 \\
SLTD-One Step &
  5.578 $\pm$ 0.050 &
  1.000 &
  36.819 $\pm$ 0.23 &
  0.412 &
  9.671 $\pm$ 0.129&
  0.531 &
  4.588 $\pm$ 0.178 &
  0.337 \\ \midrule
\begin{tabular}[c]{@{}l@{}}SLTD  (K=1)\end{tabular} &
  8.011 $\pm$ 0.025 &
  0.511 &
  36.678 $\pm$ 0.289 &
  0.297 &
  9.311 $\pm$ 0.162 &
  0.517 &
  6.492 $\pm$ 0.388 &
  0.372 \\
\begin{tabular}[c]{@{}l@{}}SLTD-Stat. (K=1)\end{tabular} &
  5.575 $\pm$ 0.036 &
  1.000 &
  34.320 $\pm$ 0.175 &
  1.000 &
  8.659 $\pm$ 0.027 &
  0.571 &
  3.662 $\pm$ 0.059 & 
  0.263\\
\begin{tabular}[c]{@{}l@{}}SLTD-One Step (K=1)\end{tabular} &
  5.595 $\pm$ 0.038 &
  1.000 &
  36.482 $\pm$ 0.284 &
  0.328 &
  8.640 $\pm$ 0.104 &
  0.318  &
  3.959 $\pm$ 0.130 &
  0.387 \\ \midrule
\shaping &
  3.044 $\pm$ 0.023 &
  0.512 &
  13.273 $\pm$ 0.126 &
  0.000 &
  N/A & 
  N/A &
  N/A &
  N/A \\
\mozannar &
  6.369 $\pm$ 2.226 &
  0.468 &
  -15.155 $\pm$ 0.675 &
  0.010 &
  3.820 $\pm$ 0.231 &
  0.291 &
  4.726 $\pm$ 0.295 &
  0.416 \\
\madras &
  5.731 $\pm$ 0.223 &
  1.000 &
  35.388 $\pm$ 0.475 &
  0.332 &
  3.330 $\pm$ 0.481 &
  0.400 &
  4.972 $\pm 0.125$ &
  0.360 \\ \midrule
$\pi_{\text{tar}}$ &
  -1.80 $\pm$ 0.071 &
  N/A &
  13.202 $\pm$ 0.162 &
  N/A &
  5.837 $\pm$ 0.171&
  N/A &
  3.292 $\pm$ 0.274 &
  N/A \\
$\pi_{0}$ &
  5.485 $\pm$ 0.035 &
  N/A &
  34.241 $\pm$ 0.151 &
  N/A &
  14.124 $\pm$ 0.592 &
  N/A &
  5.629 $\pm$ 0.159 &
  N/A \\ \bottomrule
\end{tabular}
\end{adjustbox}
\caption{Expected Value for \sltd compared with baselines. The table shows the value of using $\pi_{\text{tar}}$ with our deferral method \sltd, its variants, and other baselines. Higher values indicate better performance. Note that none of the values shown are cost-penalized, i.e. all values are true values within the system, irrespective of the deferral cost. Results show the best performing policy after tuning the respective hyperparameters (see Appendix~\ref{app:hyperparams}). We further show how often each method defers to the expert policy (lower is better). $K=1$ variants of \sltd indicate ablations of \sltd that ignore modeling uncertainty. \sltd outperforms baselines in for all datasets with fewer (though not the lowest) deferral frequency.}
\label{tab:value_results}
\end{table*}

\paragraph{Synthetic Data.}
In this synthetic simulation, the region of deferral is known apriori by careful design of $\pi_{\text{tar}}$. This environment has $8$ discrete states and binary actions $\{a_0, a_1\}$. All samples start at state $0$ and progress toward a sink state $7$. The episode length is $15$. State $6$ has low reward ($-5$) while all other states have a reward of $+1$. The initial dynamics are set up such that action $a_0$ reduces the probability of landing in stage $6$, and action $a_1$ increases the probability of reaching state $6$. $\pi_{\text{tar}}$ increases the chances to reach state $6$ unfavourably by taking action $a_1$ in states $2,3,4$ when $t < 5$ or $t > 12$. We expect to defer in states $2,3,4$ even though rewards are favorable, if a method is pre-emptive. When $5 \leq t \leq 12$, the dynamics flip such that $a_0$ becomes an unfavourable action that increases the probability of landing in $6$, while $a_1$ reduces this probability. Here, $\pi_{\text{tar}}$ again increases the chances of landing in $6$, by taking $a_0$ more often in states $2, 3, 4$. By flipping the better action to $a_0$ in this region, it becomes crucial to \emph{estimate the dynamics} over  predicting the best action. 
The dynamics are non-stationary and the probability of landing in state $6$ progressively increases when $5 \leq t \leq 12$. 


\begin{figure*}[t]
    \centering
    \begin{minipage}{\textwidth}
    \centering
    \includegraphics[scale=0.16]{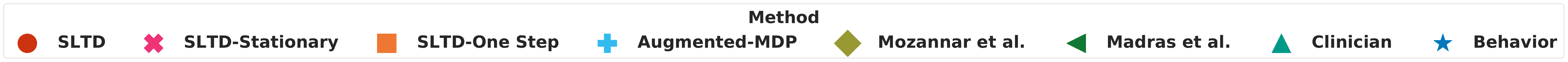}
    \end{minipage}
    \begin{minipage}{.22\textwidth}
        \centering
        \includegraphics[scale=0.092]{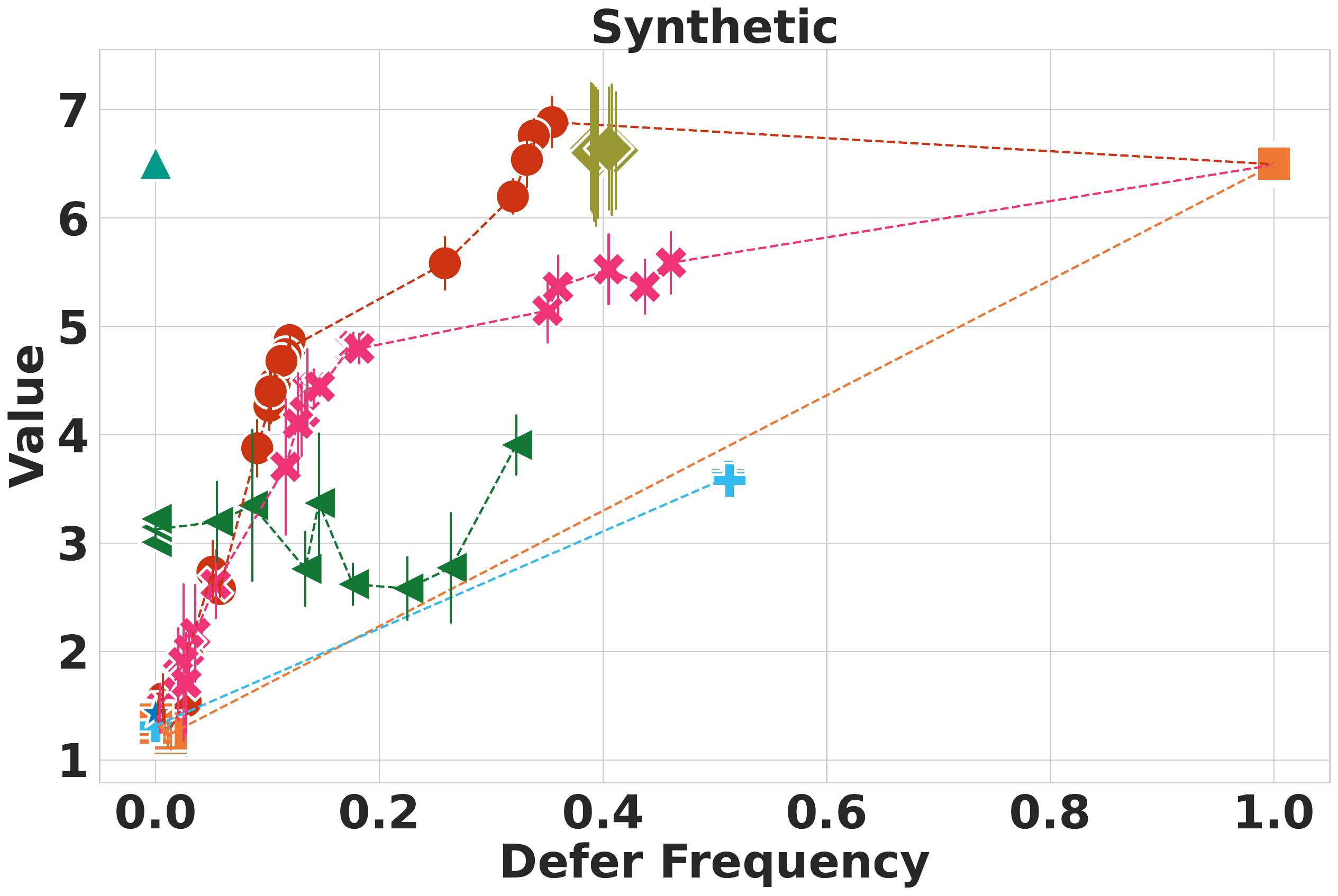}
        \label{fig:budget_disc}
    \end{minipage}%
    \begin{minipage}{0.23\textwidth}
        \centering
        \includegraphics[scale=0.092]{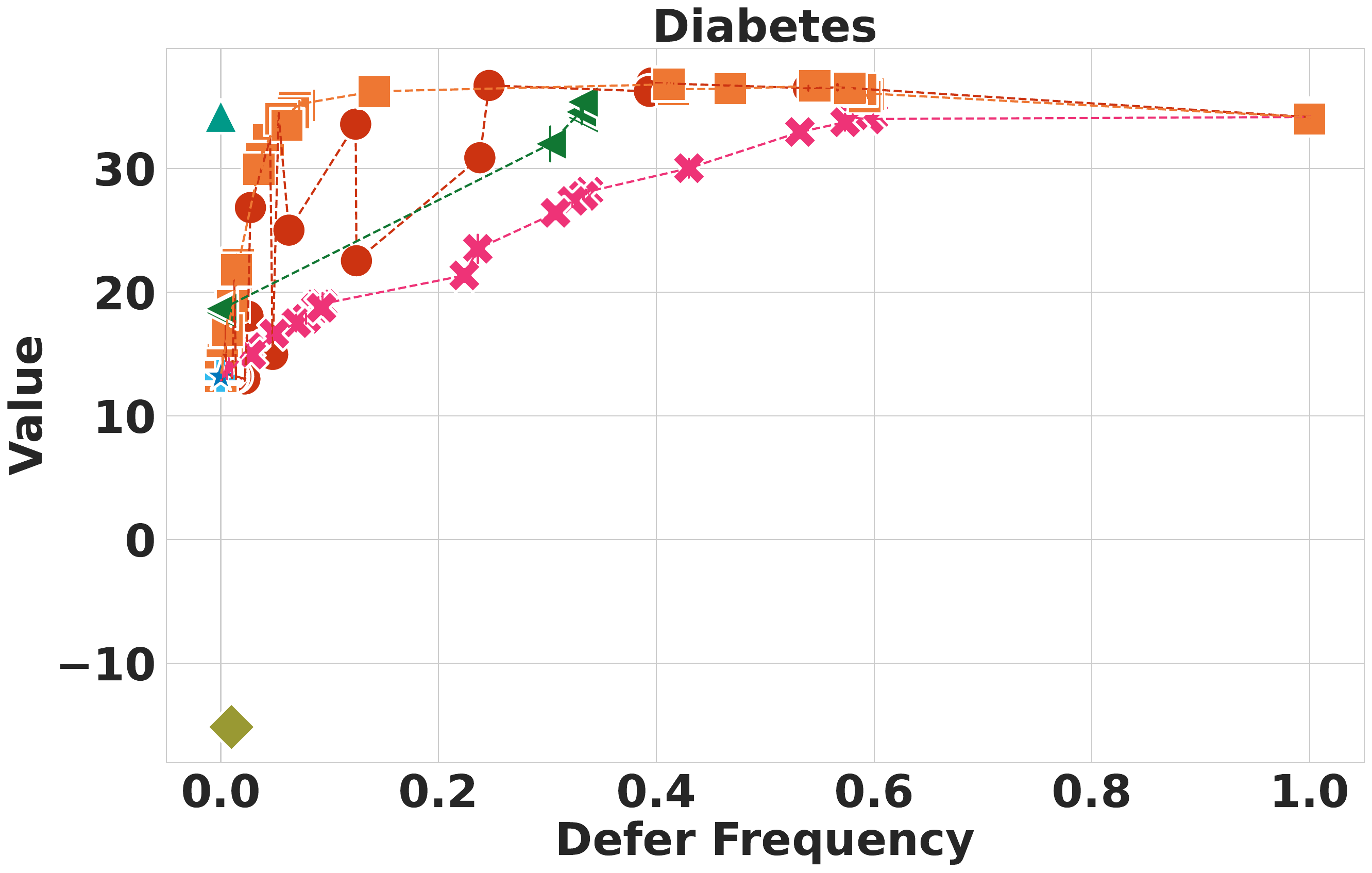}
        \label{fig:budget_sep_diab}
    \end{minipage}
    \begin{minipage}{0.23\textwidth}
        \centering
        \includegraphics[scale=0.20]{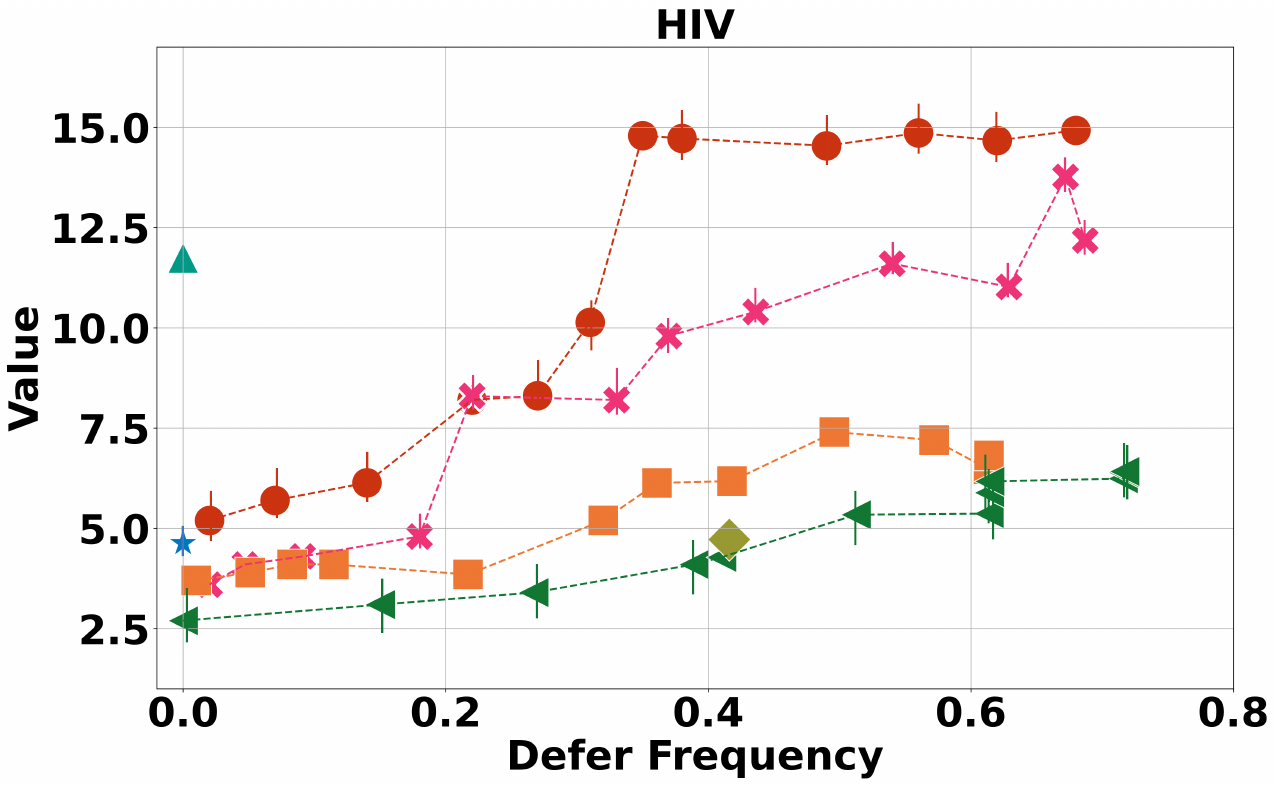}
        \label{fig:budget_sep_hiv1}
    \end{minipage}
    \begin{minipage}{0.23\textwidth}
        \centering
        \includegraphics[scale=0.19]{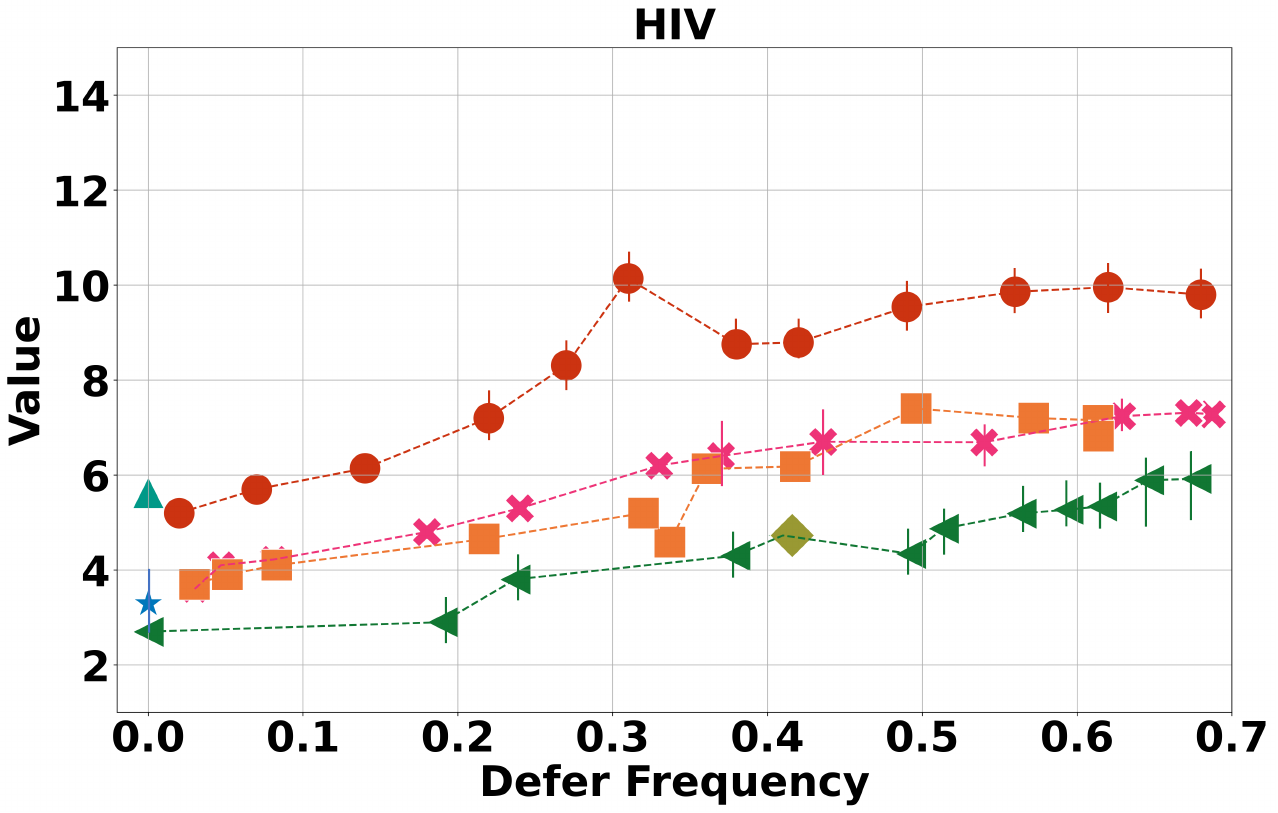}
        \label{fig:budget_sep_hiv2}
    \end{minipage}
    \caption{Trade-off of deferral frequency and value attained. Plot shows expected value (higher is better) for a sweep over deferral cost (and any other method specific hyperparamters) in relation to the deferral frequency (lower is better). \sltd achieves the best trade-off and better value by deferring pre-emptively. The stationary variant defers significantly more to achieve better outcomes. The One-Step variant cannot improve over $\pi_{\text{tar}}$ in MDPs where effect of interventions are not myopic (Synthetic and HIV) and achieve good performance by virtue of modeling dynamics in Diabetes, compared to existing learning-to-defer baselines \mozannar and \madras (which are myopic). \shaping is unable to achieve good performance even at upto 50\% deferral and unable to improve value in Diabetes indicating the benefits of deferring by explicitly quantifying the impact of delaying deferral as in \sltd over \shaping's Value Iteration method.}
    \label{fig:lineplot_all}
\end{figure*}
\paragraph{Real-world simulator: Diabetes Data.} We use an open-source implementation of the FDA approved Type-1 Diabetes Mellitus simulator (T1DMS) for modelling treatment of Type-1 diabetes. We sample $10$ adolescent patient trajectories (episodes) over $24$ hours (aggregated at $15$ minute intervals). Glucose levels are discretized into $13$ states. Combination interventions of insulin and bolus are discretized to generate a total of $25$ actions. We introduce non-stationarity in each episode by increasingly changing the adolescent patient properties to an alternative patient. This significantly affects the utility of the initial target policy which is learned on the dynamics of the original patient, thus necessitating deferral as the patient properties change over time. The non-stationary behavior policy $\pi_{\text{tar}}$ for this task is estimated using Q-learning. We defer to a clinician policy, here simulated by learning an epsilon-greedy version of a policy learned using Q-learning under (estimated) non-stationary dynamics on the target data. 

\paragraph{Real-world: HIV Data.}We identified individuals between 18-72 years of age from the EuResist database \citep{zazzi2012predicting} comprising of genotype, phenotype and clinical information of over 65,000 individuals in response to antiretroviral therapy administered between 1983-2018.  We focus on a subset of $32,960$ patients' 
  genotype, treatment response,  CD$4+$ and viral load measurements, gender, age, risk group, number of past treatments collected over on average $14$ years (aggregated at 4-6 month intervals). Our action space consists of the $25$ most frequently occurring drug combinations, while our state space consists of $100$ continuous states of cell counts and viral loads. Since the virus evolves in response to drug pressure, the problem is inherently non-stationary. For our first case study, we investigate whether deferring to a second line therapy as proposed by standard medical guidelines \citep{saag2020antiretroviral} in response to potential drug resistance improves long-term outcomes. 
 The non-stationary behaviour policy is the first line therapy estimated using Q-learning.  For our second case study, the non-stationary behaviour policy corresponds to a first line therapy typically used for treating patients of subtype C. We then examine whether deferring to a first line therapy, given by clinical collaborators, for patients of subtype M (due to potential drug resistance) improves long-term outcomes. 

{\bf{Baselines.}} We compare to the following baselines. 

\paragraph{{\mozannar}}~\citep{mozannar2020consistent}: This is a supervised method using a consistent loss function to learn-to-defer. It learns an augmented regressor to defer or recommend  treatment myopically (independently at every time-step). When the model defers, clinician policy is used. 

\paragraph{\madras} \citep{madras2017predict}: This is an alternative supervised learning-to-defer method. 
This baseline learns separate regressors to defer and  recommend treatments. We modify it to use $\pi_{tar}$ to recommend and learn the rejection function to defer to $\pi_0$.

{\bf{Augmented-MDP}}:  A conceptual contribution of \sltd is to defer by comparing outcomes by delaying deferral with some knowledge of the expert policy. The deferral action itself is considered to augment the MDP action-space. We explore a baseline that uses Value Iteration in this augmented MDP. Comparing with this baseline helps evaluate the utility of deferring based on outcomes on delayed versus immediate deferral. This baseline will defer permanently to the expert, and knowledge of an expert policy is not assumed. 
 This augmented MDP has action-space is $\cA \cup \perp$, an augmented state-space $\cS \cup s_{defer}$ ($s_{defer}$ is the deferred state), and defers based on the cost $c$. 
 This baseline models non-stationary dynamics, and is designed for sequential settings. However since this method defers permanently to the expert, it incurs a larger deferral cost. In our experiments all values are plotted \emph{without} the cost to reflect actual environment outcomes. 

{\bf{SLTD-Stationary}}: To assess the impact of mis-specifying the non-stationarity, we compare to a variant of \sltd that assumes the dynamics (and rewards) are stationary  
while allowing the method the flexibility of learning a non-stationary deferral policy.

{\bf{\sltd}-One Step}: We compare to a myopic version of \sltd that defers based on the immediate reward. The key difference with the myopic \madras, \mozannar baselines is that \sltd-One Step models the dynamics and the uncertainty on the immediate reward.  
Thus, this baseline accounts for future deferrals while deferring myopically. 

{\bf{Ablations for Uncertainty Modeling}}: For all \sltd variants, we evaluate the utility of accounting for different sources of uncertainty, more specifically the \emph{modeling uncertainty} to estimate the probability of improving outcomes via delayed deferral. In \sltd, \emph{modeling uncertainty} is accounted for by sampling multiple ($K$) MDPs (Equation~\ref{eq:G_est}) from the posterior dynamics distribution, over which our outcomes are averaged. Higher variability across $K$ indicates higher modeling uncertainty. Hence, in Equation~\ref{eq:G_est}, using $K=1$ assumes a perfect estimate of the dynamics model and only accounts for the irreducible stochasticity of the environment. A larger $K$ accounts for potential variability in estimation (original \sltd formulation). If our modeling uncertainty in the environment is indeed large, we anticipate choice of $K$ to have a larger effect on \sltd's performance. 
Modeling uncertainty can be large when there is insufficient data to fit the target function class of the dynamics. 

\section{Results}


\paragraph{Optimizing for long-term outcomes learns qualitatively different deferral policies.} Our deferral policy is a non-stationary stochastic function $\tilde{g}_{\pi_{\text{tar}}}(s,t)$ which we threshold. Visualizing $\tilde{g}_{\pi_{\text{tar}}}(s,t)$ enables us to understand the utility of various modeling choices of \sltd.  Figure~\ref{fig:def_policy_plot} shows the histograms of $\tilde{g}_{\pi_{\text{tar}}}$ for \sltd and its Stationary and One-Step variant when the cost $c=0$ for \Synthetic. Visualizing without deferral cost allows us to see the how adaptive \sltd is without a penalty. Each row corresponds to a method; x-axis corresponds to time over the horizon $T$. Each box in a row corresponds to one time point. For a fixed $t$, $\tilde{g}_{\pi_{\text{tar}}}$ is a stochastic function of the states, shown as a histogram. 

Yellow shaded region indicates the state space where $\pi_{\text{tar}}$ takes unfavorable actions. Over time, the dynamics change so that the favorable action flips $5\leq t \leq 12$ and the stochasticity in the dynamics increases requiring more frequent deferrals. Deferring in the yellow region is desirable to pre-emptively avoid landing in state $6$. 
\sltd is highly adaptive, and pre-emptively defers in the yellow region. As the stochasticity increases,  probability of deferrals appropriately increases. The stationary variant  significantly under-estimates the need to defer in states $2,3,4$ when $t<5$. 
It is only able to pre-emptively defer in regions where the average stochasticity of the estimated dynamics aligns with the environment. The One-Step variant defers only in state $6$ and is therefore not-preemptive. \shaping (red vertical line) deterministically defers in state $6$. Thus deferring based on probability of improved outcomes of immediate and delayed deferrals is desirable over alternatives 
(see also Figures~\ref{fig:diabetes_policies}, \ref{fig:HIV_policies}, and \ref{fig:HIV2_policies} in  Appendix). 

\paragraph{\sltd improves long-term outcomes.}

Table~\ref{tab:value_results} shows the value (higher is better) for all baselines corresponding to best performing parameters. For all datasets,  we see a significant benefit from \sltd. 
\shaping baseline is not preemptive in spite of modeling the non-stationary dynamics. Hence,  deferring by comparing outcomes of delayed deferral is better alternatives, specifically Value Iteration on the Augmented MDP.  
Additional benefits of modeling the dynamics is clear from improved performance of all \sltd variants compared to the myopic baselines. However, mis-specification  of  dynamics  (SLTD-Stat.) results in worse performance. SLTD-Stat. defers more often to achieve comparable performance. Figure~\ref{fig:lineplot_all} demonstrates this trade-off is general, for all choices of deferral costs and other parameters. The x-axis corresponds to deferral frequency (lower is better) and y-axis the value attained (higher is better). \sltd achieves the best trade-off. Further pre-emptive deferral allows \sltd to reach higher value than the expert policy itself. \sltd-One Step only relies on immediate rewards  failing to improve long-term outcomes for \synthetic and HIV. However as long as we model the dynamics appropriately, even myopic deferral using \sltd One Step is beneficial for Diabetes compared to \mozannar, 
\madras. This is possible when effect of interventions are observed myopically, as is the case in Diabetes data since modeling the dynamics and impact of future deferral is beneficial to characterize. 
 \madras, \mozannar baselines are unable to maximize long-term rewards. \madras performs well on Diabetes data suggesting optimal actions don't significantly deviate in target data and that its design of training a rejection function worked better than  the loss function design of  \mozannar 

\begin{table}[!htbp]
\centering
\begin{adjustbox}{max width=0.45\textwidth}
\begin{tabular}{@{}lcccc@{}}
\toprule
 &
  \begin{tabular}[c]{@{}c@{}}Defer Time\\ $t_d$\end{tabular} &
  \begin{tabular}[c]{@{}c@{}}Total \\ Uncertainty\end{tabular} &
  \begin{tabular}[c]{@{}c@{}}Modeling \\ Uncertainty\end{tabular} &
  \begin{tabular}[c]{@{}c@{}}Mean\\ Outcome\end{tabular} \\ \midrule
\Synthetic &
  3 &
  26.190 &
  0.233 &
  3.42 \\
Diabetes &
  3 &
  3418.17 &
  34.160 &
  73.669 \\ \bottomrule
\end{tabular}
\end{adjustbox}
\caption{Interpreting first time of deferral for a sample trajectory. Modeling uncertainty remains low in all cases whereas in comparison, total variance is high. This indicates irreducible stochasticity of the dynamics is the primary source of uncertainty. Additional results are in Appendix.}
\label{tab:interpret_decomp}
\end{table}

\paragraph{Ablations for uncertainty modeling.} 
We study the utility of accounting for modeling uncertainty in our framework. As described in Section~\ref{sec:unc_decomp}, multiple sources of propagated uncertainty contribute to  variability in estimated outcomes. 
Modeling uncertainty is crucial to account for in a model-based framework. Here we evaluate the impact of not accounting for this uncertainty on \sltd's performance. 

If modeling uncertainty is high, variability of the sampled MDPs used to estimate Equation~\ref{eq:G_est} will be higher. Evaluating for $K=1$, will evaluate the impact of ignoring this uncertainty.  
In Table~\ref{tab:value_results} (see also Figure~\ref{fig:uncertainty_ablations} in Appendix), we demonstrate the results with $K=1$ for all \sltd variants. 
We do not observe significant differences for \Synthetic and Diabetes indicating that our modeling uncertainty is low in these data. The difference is higher in HIV suggesting the importance of accounting for this uncertainty for real-world HIV data. 
Such analysis is crucial to understanding whether our modeling assumptions are reasonable.

\paragraph{Decomposing uncertainty in \sltd can help interpret deferral.} 
Conveying 
the type of uncertainty to a domain expert can help identify the dominant source of uncertainty that resulted in a deferral to their own standard practice (expert policy). Table~\ref{tab:interpret_decomp} shows this decomposition for one  time point for discrete data. In each case, the modeling  uncertainty is a small fraction of the total uncertainty. This suggests that the systematic non-stationarity is the dominant source of uncertainty which generally cannot be reduced by collecting data and may require careful interventions beyond standard policy. Knowledge of the amount of model uncertainty can enable users to further improve decision-making through data collection or improving model assumptions. 

\section{Discussion}\label{sec:discussion}
We proposed \sltd, a learning-to-defer framework for sequential settings using offline model-based RL. 
We learn a deferral policy by quantifying the impact of delaying deferral to the future. \sltd is able to defer based on long-term outcomes, and  learns a pre-emptive deferral policy. 
Further, we emphasize a model-based RL method that captures the dynamics of the environment, particularly non-stationarity. Modeling non-stationarity of the environment allows to defer adaptively. Misspecifying non-stationarity leads to significantly more deferrals to improve long-term outcomes. 
 We demonstrate that existing learning-to-defer frameworks are myopic. That is, these  methods do not learn a pre-emptive policy even in  sequential settings as they focus on immediate consequences of actions. 
We further demonstrate the utility of accounting for all potential sources of stochasticity to quantify the impact of delayed deferral. Explicit characterization of probability of improving outcomes is beneficial to prevent over-estimation of benefits of delaying deferral. 
We further interpret deferral decisions of \sltd by decomposing the \emph{long-term} propagated uncertainty.  
While quantifying the uncertainty is useful, 
modeling uncertainty through  non-stationarity is costly. Developing a model-free  framework is left for future work.  \sltd can account for some deviations to the expert policy, though significant deviations could be modeled as a human-in-the-loop and is left to future work. 

{\bf{Ethical considerations.}} 
\sltd is a technical proof-of-concept to defer to an expert by accounting for long-term effects, assuming that the expert is better at increasing value over the current policy in certain regions. In practice, an expert policy may not be bias free. Thus, deferral may result in biased decisions if the expert is biased. While we are not focused on addressing bias, 
 exposing uncertainties may encourage expert introspection. Nonetheless, deferring  is better when an  automated decision may be harmful.

\bibliographystyle{plainnat}
\bibliography{ltd-transfer}

\clearpage
\newpage
\section{Appendix}

\subsection{Bayesian RL for Dynamics and Rewards Estimation.}\label{app:posterior}

We assume that the dynamics are governed by a sequence of MDPs to allow for non-stationarity. The dynamics of each MDP in the sequence, denoted by $\cP_t$ is estimated using batch data. Our batch data $\cD^*$ can be split into tuples indexed by the time $t$ as: $\cD^* \triangleq \{\cD_t^*\}_{t=1}^T$. To infer $\cP_t$, batch samples $\cD_t^*$ will be used. 

{\bf{Discrete State.}} We first describe the procedure for discrete states. The distribution characterizing the dynamics $\cP_t$ are parameterized by $\theta_t \triangleq \{\theta_t(s, a)\}_{(s,a)}$ which is a tuple over all state-action pairs. In particular, we use a Bayesian framework to estimate the parameters of the dynamics distribution $\cP_{\theta, t}$. For discrete data, $\cP_{\theta, t} \triangleq p(\cD_t^* \vert \theta_t)$ are modeled as Multinomial distributions for each state-action pair $(s,a)$. That is, this parameter $\theta(s,a) \in \Delta^{|\cS|-1}$ lies in the probability simplex of dimension $|\cS|-1$ and $p(s' \vert s, a) = \theta_{s'}(s, a)$. The prior $p(\theta_t(s, a))$ is assumed to follow a  Dirichlet Distribution.  The Dirichlet prior distribution is parameterized by $\alpha \in \mathbb{R}^{|\cS|-1}$ and is given by, $$p(\theta) = \frac{1}{B(\alpha)}\prod_{i=1}^{|\cS|}\theta_i^{\alpha_i-1}$$ where $B(\alpha)$ is the multivariate Beta function.

The posterior distribution of $\theta(s,a)$ given samples $\cD_t^*$ is itself a Dirichlet distributed random variable (since Dirichlet distribution is a conjugate prior for the Multinomial likelihood  distribution), with parameters $\alpha_t'$,
\begin{equation}
    p(\theta_t(s,a) | \cD_t^*) = \text{Dirichlet}(\alpha_t')
\end{equation}
where $\alpha'_{s',t} = \alpha_{s',t} + \sum_{s'' \in \cD^*_t} \mathbf{1}(s'' == s')$. 

Note that we also account for uncertainty over rewards by estimating posteriors via Bayesian inference, as in the case of the dynamics. For discrete rewards, Dirichlet priors are used analogously. 

{\bf{Continuous State.}} For continuous states, the dynamics are assumed normally distributed. The parameters of the normal distribution are assumed to have a normal-gamma prior. That is, Let $\mu_t(s,a), \tau_t(s,a)$ be the mean, and precision of the parameter describing the dynamics as a function of the states and actions. The normal-gamma prior is given as follows:
\begin{equation}
\begin{aligned}
    \mu_t(s,a) \vert \tau(s,a) &\sim \cN(\mu_0, n_0\tau(s,a)) \\
    \tau(s,a) &\sim \text{Gamma}(\eta, \beta) \\
\end{aligned}
 \end{equation}
 
 Note that we use the same prior distribution for all state-action pairs, though custom priors may be used based on domain-knowledge. The posterior distributions after observing data samples $\cD^*_t$, specifically $\bs'$ over $\mu_t(s,a)$ are given by a Gaussian with the following parameters:
 \begin{equation}
     \mu(s,a) | \tau(s,a),\bs' \sim \cN \Bigl(\mu_t',  \tau_t' \Bigr)
 \end{equation}
 \begin{equation}
     \mu_t' =  \frac{n_t\tau(s,a)}{n_t \tau(s,a) + n_0 \tau(s,a)}\bar{s'} + \frac{n_0 \tau(s,a)}{n_t\tau(s,a) + n_0 \tau(s,a)}\mu_0
 \end{equation}
 
 \begin{equation}
     \tau_t' = n_t \tau(s,a) + n_0 \tau(s,a)
 \end{equation} 
 where $\bar{s'}$ is the mean of the observations $\bs'$ and $n_t$ are the number of observations for state-action $s,a$ observed at time $t$. The posterior distribution over the precision $\tau(s, a)$ is given by,
 $$\tau(s, a) \vert \bs' \sim \text{Gamma}(\eta', \beta')$$ where,
 
 $$\eta' = \eta_0 + \frac{n_t}{2}, \beta' = \beta_0 + \frac{1}{2}\left(n_t  \sum_{i}(s_i' - \bar{s})^2 + \frac{n_tn_0 (\bar{s}' - \mu_0)^2}{2(n_t+n_0)}\right)$$

\subsection{Decomposing Propagated Uncertainty}\label{app:uncertainty_decomp}
We describe how uncertainty of the long-term outcome $\mathbb{E}[r_{T}|s_{t_{d}},, \mu_{t_d}]$ at deferral time $t_d$, when the agent is in state $s_{t_d}$ can be decomposed into modeling/epistemic uncertainty and irreducible/aleatoric uncertainty in the following. Once we defer, we sample actions from $\pi_0$ at time $t'=t_d$ and $\pi_{\text{mix}}$ for $t'> t_d$ where the mixture probability is determined by $g_{\pi_{\text{tar}}}$ for future deferrals. The expected long-term outcome is given by:
{\small{
\begin{align*}
    \begin{split}
        &\mathbb{E}[r_{T}|s_{t_{d}}, \mu_{t_d}] = 
        \int_{s_{t_{d}+1}}^{s_{T}} \int_{a_{t_{d}}}^{a_{T}} \int_{\mu_{t_{d}+1}}^{\mu_{T}} \int_{\theta_{t_d}}^T r(s_T, a_T) \times 
        \prod_{t'={t_{d}+1}}^T p_{t'}(s_{t'} | \mu_{t'}) p_{t'}(\mu_{t'} | \theta_t'(s_{t'}, a_{t'})) 
        \pi_{t'}(a_{t'}| s_{t'}) p_{t'}({\theta_{t'}}| \cD) d\bs d\ba  d\bmu d{\btheta} 
    \end{split}
\end{align*}
}}
Integrands are written in short-hand: $\bs = \{s_{t_d+1}, s_{t_d+2}, \cdots, s_{T}\}$ (analogously for other quantities). As suggested before, $\pi_{t_d} = \pi_{0(t_d),\text{mix}(t_d+)}$ to account for future deferrals. Here, we denote the posterior MDP samples for any state action pair by $\mu_{t}$. 
 The variability in these samples capture modeling uncertainty. The dynamics parameters are denoted by $\theta_t$ for each state-action pair. We sample from posterior distribution $p(\theta_{t'} | \cD^*)$, followed by sampling the MDPs $\mu_{t} \sim p(\mu_{t'} | \theta_t'(s_{t'}, a_{t'}))$. We maintain one estimate of parameter $\theta_{t'}$ and sample $K$ MDPs $\mu_{t'}$ from this distribution. That is, $p(\theta_{t'}| \cD ) = \delta_{\theta_{t'}}$ which is a delta function centered at $\theta_{t'} \, \forall \, t' \in \{0, 1, 2, \cdots, T\}$:
 
 {\small{
\begin{align*}
    \begin{split}
        &\mathbb{E}[r_{T}|s_{t_{d}}, \mu_{t_d}] = 
        \int_{s_{t_{d}+1}}^{s_{T}} \int_{a_{t_{d}}}^{a_{T}} \int_{\mu_{t_{d}+1}}^{\mu_{T}} \int_{\theta_{t_d}}^T r(s_T, a_T) \times 
        \prod_{t'={t_{d}+1}}^T p_{t'}(s_{t'} | \mu_{t'}) p_{t'}(\mu_{t'} | \theta_t'(s_{t'}, a_{t'})) 
        \pi_{t'}(a_{t'}| s_{t'}) \delta_{\theta_{t'}} d\bs d\ba  d\bmu 
    \end{split}
\end{align*}
}}
Thus, the epistemic uncertainty we capture is due to the uncertainty over dynamics under fixed parameters. High variability in sampling $\mu_{t'}$ indicate the current state $s_{t'}$ (and action) are out-of-distribution. The total uncertainty can now be decomposed using the law of total variance:
\begin{equation*}
    \text{Var}(Y) = \mathbb{E}[\text{Var}(Y|X)] + \text{Var}(\mathbb{E}[Y|X])
\end{equation*}
Applying this to our target outcome, we have:

\begin{align}\label{eq:unc_decom_app}
    \begin{split}
     \underbrace{\text{Var}(r_{T} | s_{t_d}, \cD)}_{\text{Total Uncertainty}} = &\underbrace{\mathbb{E}_{\mu_{t_d} \sim p(\mu_{t_d}| \cD)}\big[\text{Var} (r_T| \mu_{t_d}, s_{t_d}, \cD)\big]}_{\text{Irreducible/ Aleatoric Uncertainty}}  + 
 \underbrace{\text{Var}_{\mu_{t_d} \sim p(\mu_{t_d}| \cD)} \big( \mathbb{E}[r_{T}| \mu_{t_d}, s_{t_d}, \cD]\big)}_{\text{Epistemic/Modeling  Uncertainty}}
    \end{split}
\end{align}

The second term  is the variance \emph{conditioned} on knowledge of the model $\mu_{t_d}$, therefore marginalizing only over current aleatoric uncertainty and future total uncertainty (i.e. over future $\mu_{t'}$, future deferral, and reward). 
 This is the \emph{propagated uncertainty due to modeling uncertainty at $t_d$}, and can be reduced by data collection. The first term averages over the variance due to $\mu_{t_d}$ and captures \emph{propagated  uncertainty to due to aleatoric uncertainty at $t_d$}, which can only be reduced by careful interventions at $t_d$. We estimate these using Monte-Carlo sampling. High \emph{propagated epistemic uncertainty} conveys that the current uncertainty of model prediction (of the dynamics) is high but could be improved if additional data could be collected. High \emph{propagated aleatoric uncertainty} indicates high variability in the patient's dynamics that can only be  with careful interventions and is otherwise not manageable. Based on the communicated uncertainty, the clinician may choose to deviate from their usual practice for rare cases with high epistemic uncertainty and instead consult multiple experts or attempt experimental treatments. 

\subsection{Datasets}

{\bf{Discrete Toy Data.}} All state and action spaces are discrete. True dynamics of the environment are known. This environment has $8$ discrete states and binary actions $\{a_0, a_1\}$. All samples start at state $0$ and progress toward a sink state $7$. The episode length is $15$. State $6$ has low reward ($-5$) while all other states have a reward of $+1$. The initial dynamics are set up such that action $a_0$ reduces the probability of landing in stage $6$, and action $a_1$ increases the probability of reaching state $6$. $\pi_{\text{tar}}$ increases the chances to reach state $6$ unfavourably by taking action $a_1$ in states $2,3,4$ when $t < 5$ or $t > 12$. We expect to defer in states $2,3,4$ even though rewards are favorable, if a method is pre-emptive. When $5 \leq t \leq 12$, the dynamics flip such that $a_0$ becomes an unfavourable action that increases the probability of landing in $6$, while $a_1$ reduces this probability. Here, $\pi_{\text{tar}}$ again increases the chances of landing in $6$, by taking $a_0$ more often in states $2, 3, 4$. By flipping the better action to $a_0$ in this region, it becomes crucial to \emph{estimate the dynamics} over  predicting the best action. 
The dynamics are non-stationary and the probability of landing in state $6$ progressively increases when $5 \leq t \leq 12$. The reward vector for Discrete Toy data is a function of states only and is given by:

\[ r(s) = \begin{cases*}
                    1 & if  $s \neq 6$  \\
                     -5 & if $s = 6$
\end{cases*} \]%

\paragraph{Diabetes simulator.} We use an open-source implementation of the FDA approved Type-1 Diabetes Mellitus simulator (T1DMS) for modelling treatment of Type-1 diabetes. The simulator models the dynamics of an in-silico patient's blood glucose levels when consuming a meal. If the blood glucose level is either too high (hyperglycemia) or too low (hypoglycemia), this can have fatal consequences such as organ failure. As a result, a clinician must administer an insulin dosage to minimize the risk of such events. While a doctor’s initial dosage prescription is usually available, the insulin sensitivity of a patient’s internal organs changes over time, thereby introducing non-stationarity that should be accounted for. We sample $10$ adolescent patient trajectories (episodes) over $24$ hours (with measurements aggregated at $15$ minute intervals). Glucose levels are discretized into $13$ states according to ranges suggested in the simulator. Further, insulin and bolus intervention combinations are discretized to generate a total of $25$ actions. The non-stationary characteristics of this publicly available simulator allow us to thoroughly evaluate all aspects of SLTD under the true dynamics of the simulator. The reward function for Diabetes data is stationary and defined in the simulator. It is a function of the state and is defined as the change in risk due to change in blood glucose level between the last two measurements. Discretization of Glucose levels is provided in Table~\ref{tab:diabetes_bg} and discretization of interventions is summarized in Table~\ref{tab:bolus} (bolus) and Table~\ref{tab:insulin} (insulin). The discrete combinations of bolus and insulin are combined to generate $25$ potential actions.  

\begin{table}[!htbp]
\centering
\begin{tabular}{@{}ll@{}}
\toprule
\begin{tabular}[c]{@{}l@{}}Blood Glucose\\ (mg/dL) - BG\end{tabular} & Discrete state \\ \midrule
$0< \text{BG} \leq 29.2$    & $0$ \\
$29.2 < \text{BG} \leq 58.5$   & $1$ \\
$58.5 < \text{BG} \leq 87.8$  & $2$ \\
$87.8 < \text{BG} \leq 116.9$ & $3$ \\
$116.9 < \text{BG} \leq 146.2$ & $4$ \\
$146.2 < \text{BG} \leq 175.4$ & $5$ \\
$175.4 < \text{BG} \leq 204.6$ & $6$ \\
$204.6 < \text{BG} \leq 233.9$ & $7$ \\
$233.9 < \text{BG} \leq 263.1$ & $8$ \\
$263.1 < \text{BG} \leq 292.3$ & $9$ \\
$292.3 < \text{BG} \leq 321.6$ & $10$ \\
$350.8 < \text{BG} \leq 365.4$ & $11$ \\
$> 365.4$  & $12$ \\
 \bottomrule
\end{tabular}
\caption{Discretization of Blood Glucose for the Diabetes simulator}
\label{tab:diabetes_bg}
\end{table}
We introduce non-stationarity within each episode by increasingly changing the adolescent patient properties to an alternative patient over the episode. This is different from the setting  of~\cite{chandak2020optimizing} where non-stationarity is indeed across episodes. Thus our setting is more challenging. This significantly affects the utility of the initial target policy necessitating deferral as the patient properties change over the course of the day. The non-stationary clinician/behavior policy is estimated using Q-learning. We use an epsilon-greedy version of such a policy. 

\begin{table}[!htbp]
\centering
\begin{tabular}{@{}lll@{}}
\toprule
\begin{tabular}[c]{@{}l@{}}Bolus\\ (g/min)\end{tabular}  \\ \midrule
0.00  - 18.6   \\
18.6 - 37.2  \\
37.2 - 55.8   \\
55.8 - 74.4  \\
74.4 -   \\ \bottomrule
\end{tabular}
\caption{Discretization of bolus for creating combination treatments}
\label{tab:bolus}
\end{table}

\begin{table}[!htbp]
\centering
\begin{tabular}{@{}lll@{}}
\toprule
 \begin{tabular}[c]{@{}l@{}}Insulin\\ (U/min)\end{tabular} \\  \midrule
0.00  - 2.5   \\
2.5 - 5.5  \\
5.5 - 8.5   \\
8.5 - 11.5  \\
11.5 -   \\
\bottomrule
\end{tabular}
\caption{Discretization of insulin for creating combination treatments}
\label{tab:insulin}
\end{table}

Incorporating non-stationarity into the simulator:
We use the ``Navigator'' sensor to generate blood-glucose measurements and the ``Insulet'' pump to simulate interventions. For each episode, non-stationarity is induced by modifying the patient configurations over a period of $24$ hours. This result in different dynamics over the course of the day. These configurations modify insulin sensitivity, glucose absorption and the insulin action on glucose production among other parameters. For each episode, two random adolescent patients are sampled (say `a', and `b'), over every minute the patient parameters are then sampled as a convex combination of patient `a' and patient `b' where, as we progress in time, the convex combination increasingly shifts from 0 to 1 thus changing patient parameters. Over the episode, the patient parameters increasingly look like that of patient `b' instead of `a'. The rate of change of this convex combination can be controlled and is set to $\text{cos}(t\times \text{speed} \times 0.0005) \times 0.5 + 0.5$, where $\text{speed}=5$ for our simulations. A similar policy was used by \citet{chandak2020optimizing} to induce non-stationarity. However \citet{chandak2020optimizing} do not induce non-stationarity within an episode, but across different episodes. The target policy $\pi_{\text{tar}}$ is learned on data collected from patients whose dynamics do not change over time.

\paragraph{HIV Data.}
This dataset is publicly available upon request and is a continuous state dataset. We identified individuals between 18-72 years of age from the EuResist database \citep{zazzi2012predicting} comprising of genotype, phenotype and clinical information of over 65,000 individuals in response to antiretroviral therapy administered between 1983-2018. Patients are administered combinations of different drugs to prevent drug resistance and the development of viral mutations that could potentially result in resistance. Once resistance to a particular drug occurs, it is also possible for cross-resistance to develop to similar antiretrovirals from the same class, thus limiting a patient's potential treatment options for the future.  As a result, a clinician must administer antiretrovirals to minimize the risk of such resistance, while lowering the viral load in the blood. While several therapy guidelines based on clinician expertise are available, depending on what therapies a patient has previously been administered, several new mutations may develop in response to therapy, resulting in new drug-resistant variants to emerge that potentially change over time, thereby introducing non-stationarity that should be accounted for. Viral evolution (via the development of mutations and resistance to certain drugs) across different populations has led to the emergence of different HIV strains, some of which are easier to treat than others.

We focus on $32,960$ patients' genotype, treatment response,  CD$4+$ and viral load measurements, gender, age, risk group, number of past treatments collected over on average $14$ years (aggregated at 4-6 month intervals). Our state space consists of continuous states of cell counts, viral loads and mutations. Drug combinations are discretized to produce $25$ actions of the most frequently occurring combinations. For our first case study, we investigate whether deferring to a second line therapy as proposed by standard medical guidelines \citep{saag2020antiretroviral} in response to potential drug resistance improves long-term outcomes. Here, our clinician policy corresponds to a second line course of therapy, provided by our clinical collaborators. 
 The non-stationary behaviour policy is the first line therapy estimated using Q-learning.  For our second case study, the non-stationary behaviour policy corresponds to a first line therapy typically used for treating patients of subtype C. We then examine whether deferring to a first line therapy, given by clinical collaborators, for patients of subtype M (due to potential drug resistance) improves long-term outcomes. The therapies considered for both HIV Case I and II are shown in Table \ref{tab:hiv-drugs}.
 
\begin{table}[t]
\centering
\begin{tabular}{@{}ll@{}}
\toprule
\begin{tabular}[c]{@{}l@{}}Drug Combination\end{tabular} & Discrete Action\\ \midrule
$TDF+FTC+EFV$   & $0$ \\
$TDF+FTC+ATV/r$  & $1$ \\
$3TC+AZT+LPV/r $  & $2$ \\
$TDF+FTC+DRV/r$ & $3$ \\
$TDF+FTC+LPV/r$ & $4$ \\
$3TC+ABC+LPV/r$ & $5$ \\
$3TC+TDF+d4T$ & $6$ \\
$3TC+AZT+EFV$ & $7$ \\
$3TC+ABC+EFV$ & $8$ \\
$3TC+TDF+LPV/r$ & $9$ \\
$3TC+AZT+NVP$ & $10$ \\
$3TC+ddl+EFV$ & $11$ \\
$TDF+FTC$  & $12$ \\
$3TC+d4T+LPV/r$ & $13$ \\
$3TC+DRV/r+ABC$ & $14$ \\
$EFV+3TC$ & $15$ \\
$EFV+3TC+FTC$ & $16$ \\
$DTG/3TC/TDF$ & $17$ \\
$DTG/3TC$ & $18$ \\
$AZT+TDF+LPV/r$ & $19$ \\
$d4T+NFV$ & $20$ \\
$ddl+d4T+SQV/r$  & $21$ \\
$ddl+TDF+EFV$ & $22$ \\
$AZT+ddl+NVP$ & $23$ \\
$ddl+d4T+IDV/r$ & $24$ \\
 \bottomrule
\end{tabular}
\caption{Drug combinations considered for HIV Case I and II}
\label{tab:hiv-drugs}
\end{table}

\subsection{Hyperparameters and Settings.}\label{app:hyperparams}
\begin{table}[!htbp]
\centering
\begin{tabular}{@{}lll@{}}
\toprule
Method                   & Best parameter I & Best parameter II \\ \midrule
\sltd                    & $c (0.01)$       & $\tau (0.2)$      \\
\sltd-Stationary         & $c (5.00)$       & $\tau (0.0)$      \\
\sltd-One Step           & $c(0.02)$        & $\tau(0.0)$       \\
\sltd $(K=1)$            & $c(5.00)$        & $\tau(0.0)$       \\
\sltd-Stationary $(K=1)$ & $c(0.05)$        & $\tau(0.0)$       \\
\sltd-One Step $(K=1)$   & $c (0.05)$       & $\tau (0.0)$      \\
\shaping                 & $c (0.01)$       & NA                \\
\mozannar                & $c (0.1)$        & $\alpha (0.0)$    \\
\madras                  & $c (0.01)$       & NA                \\ \bottomrule
\end{tabular}
\caption{Best parameters for all methods for \Synthetic. Corresponding values and deferral frequencies are shown in Table \ref{tab:value_results}.}
\label{tab:disc_best_params}
\end{table}

\begin{table}[!htbp]
\centering
\begin{tabular}{@{}lll@{}}
\toprule
Method                   & Best parameter I & Best parameter II \\ \midrule
\sltd                    & $c (0.01)$       & $\tau (0.4)$      \\
\sltd-Stationary         & $c (0.50)$       & $\tau (0.0)$      \\
\sltd-One Step           & $c(0.20)$        & $\tau(0.4)$       \\
\sltd $(K=1)$            & $c(0.01)$        & $\tau(0.4)$       \\
\sltd-Stationary $(K=1)$ & $c(0.01)$        & $\tau(0.0)$       \\
\sltd-One Step $(K=1)$   & $c (0.1)$        & $\tau (0.4)$      \\
\shaping                 & $c (0.02)$       & NA                \\
\mozannar                & $c (0.1)$        & $\alpha (0.0)$    \\
\madras                  & $c (1.0)$        & NA                \\ \bottomrule
\end{tabular}
\caption{Best parameters for all methods for Diabetes. Corresponding values and deferral frequencies are shown in Table \ref{tab:value_results}.}
\label{tab:diabetes_best_params}
\end{table}

\begin{table}[t]
\centering
\begin{tabular}{@{}lll@{}}
\toprule
Method                   & Best parameter I & Best parameter II \\ \midrule
\sltd                    & $c (0.50)$       & $\tau (0.4)$      \\
\sltd-Stationary         & $c (0.50)$       & $\tau (0.0)$      \\
\sltd-One Step           & $c(0.20)$        & $\tau(0.2)$       \\
\sltd $(K=1)$            & $c(0.50)$        & $\tau(0.2)$       \\
\sltd-Stationary $(K=1)$ & $c(0.50)$        & $\tau(0.0)$       \\
\sltd-One Step $(K=1)$   & $c (0.2)$        & $\tau (0.2)$      \\
\shaping                 & NA       & NA                \\
\mozannar                & $c (0.1)$        & $\alpha (0.0)$    \\
\madras                  & $c (1.0)$        & NA                \\ \bottomrule
\end{tabular}
\caption{Best parameters for all methods for HIV-Case I. Corresponding values and deferral frequencies are shown in Table \ref{tab:value_results}.}
\label{tab:hiv1_best_params}
\end{table}

\begin{table}[!htbp]
\centering
\begin{tabular}{@{}lll@{}}
\toprule
Method                   & Best parameter I & Best parameter II \\ \midrule
\sltd                    & $c(0.50)$       & $\tau (0.2)$      \\
\sltd-Stationary         & $c(0.50)$       & $\tau (0.0)$      \\
\sltd-One Step           & $c(0.50)$        & $\tau(0.2)$       \\
\sltd $(K=1)$            & $c(0.50)$        & $\tau(0.2)$       \\
\sltd-Stationary $(K=1)$ & $c(0.50)$        & $\tau(0.0)$       \\
\sltd-One Step $(K=1)$   & $c (0.5)$        & $\tau (0.2)$      \\
\shaping                 & NA       & NA                \\
\mozannar                & $c (0.5)$        & $\alpha (0.0)$    \\
\madras                  & $c (1.0)$        & NA                \\ \bottomrule
\end{tabular}
\caption{Best parameters for all methods for HIV-Case II. Corresponding values and deferral frequencies are shown in Table \ref{tab:value_results}.}
\label{tab:hiv2_best_params}
\end{table}

{\bf{\sltd and variants.}} For all \sltd (original and variants), we sweep over $c \in \{0.0, 0.01, 0.02, 0.05, 0.1, 0.2, 0.5, 1.0, 2.0, 5.0, 10.0\}$ and thresholds $\tau \in \{0.0, 0.2, 0.4, 0.6, 0.8, 1.0\}$. $K=200$ for original version and $K=1$ for ablations. We use $5$ bootstraps in each run.

{\bf{\mozannar.}} This baseline has a loss penalty parameter $\alpha$. We sweep over $\alpha \in \{0.0, 0.2 0.5, 1.0, 2.0, 5.0, 10.0\}$ and show results for the best performing $\alpha$.

{\bf{\madras, \& \shaping}} This baseline has a cost parameter $c$. We sweep over $c \in \{0.0, 0.01, 0.02, 0.05, 0.1, 0.2, 0.5, 1.0, 2.0, 5.0, 10.0\}$ and show results for the best performing $c$.

All baselines were run for $5$ random seeds and the average results are shown. For the deferral frequency versus value analysis, values corresponding to all choices of parameters of the respective methods are shown to demonstrate the generality of our conclusions.

\subsection{Evaluation.}
{\bf{Value and Deferral Frequency Evaluation.}} To evaluate all methods we collect virtual roll-outs under the true dynamics. This is possible for Discrete Toy and Diabetes datasets. For HIV data the estimate of the dynamics are obtained using maximum likelihood estimates. We average cumulative rewards over $1000$ trajectories for each method. Deferral frequency is measured as the average deferral in these trajectories. 

{\bf{Uncertainty Decomposition.}} We estimate the modeling/epistemic and irreducible/aleatoric uncertainty. This uncertainty decomposition requires the posterior estimates over the MDPs. We collect this for one sample trajectory for discrete datasets as follows. First we roll out until \sltd defers. Once we defer the first time, we simulate $10000$ trajectories. Using the empirical estimates of the varianc decomposition provided in Equation~\ref{eq:unc_decom_app}, we can estimate all sources of uncertainty.  For all baselines, the cost of deferral is constant for each time-step. Uncertainty decomposition was estimated corresponding to the best performing cost $c$ and threshold $\tau$.  

\subsection{Computation Infrastructure}
All code is implemented using Python 3.8. Discrete Toy and Diabetes experments are models were trained on a single Intel 8268 ``Cascade Lake" CPUs using minimum 12GB of memory. HIV results were trained on Intel``Ice Lake" CPUs with minimum 12GB of memory. Operating system: CentOS7. Code has also been reproduced on MacOS Monterey 12.5 (8 GB 2133 MHz LPDDR3, 2.3 GHz Dual-Core Intel Core i5 and 16 GB 3.2 GHz LPDDR4, Apple M1). Code appendix includes Anaconda package dependencies required to reproduce the results.

\subsection{Additional Results}


\begin{figure*}
    \centering
    \includegraphics[width=\textwidth]{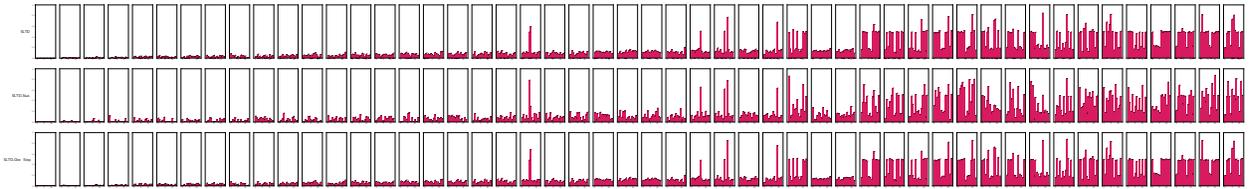}
    \caption{Learned deferral policy to Diabetes data for SLTD, SLTD-Stationary and One-Step baselines. As the dynamics shifts over time, the probability of deferring signficantly increases. Minor qualitative changes across these baselines are observed suggesting Diabetes to be largely  myopic (i.e., effect of interventions are observed in the near future). Modeling the non-stationarity remains crucial to obtain higher value (see Table~\ref{tab:value_results}).}
    \label{fig:diabetes_policies}
\end{figure*}

\begin{figure*}
    \centering
    \includegraphics[width=\textwidth]{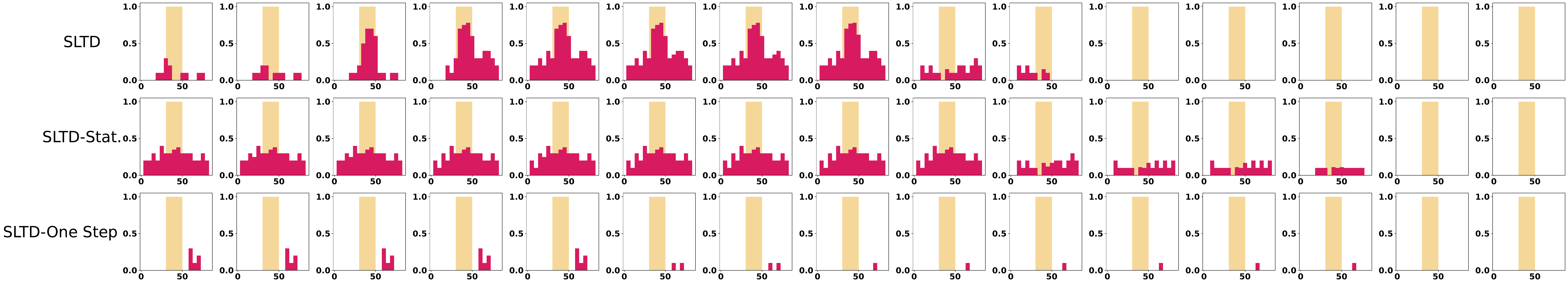}
    \caption{Learned deferral policy to HIV-I data. Shaded yellow the region of pre-emptive deferral based on clinical expertise.}
    \label{fig:HIV_policies}
\end{figure*}

\begin{figure*}
    \centering
    \includegraphics[width=\textwidth]{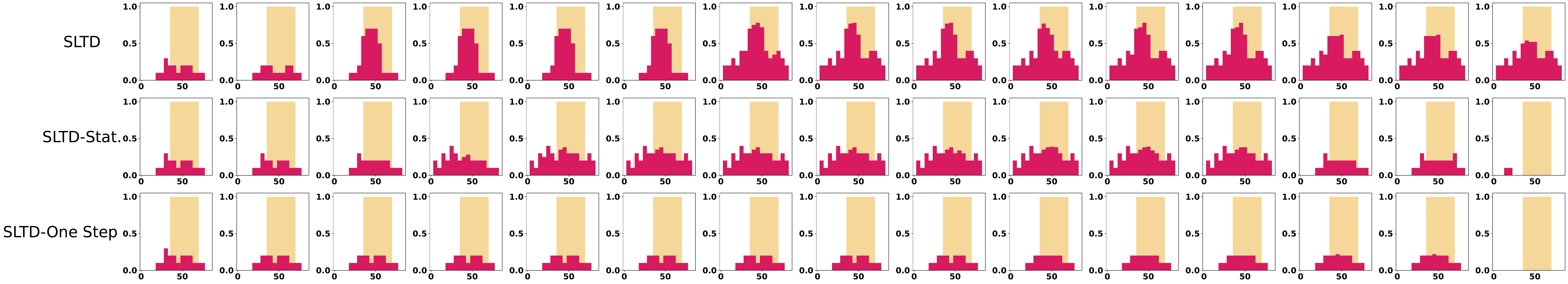}
    \caption{Learned deferral policy to HIV-II data. Shaded yellow the region of pre-emptive deferral based on clinical expertise.}
    \label{fig:HIV2_policies}
\end{figure*}

\begin{figure*}[!htbp]
    \centering
    \begin{minipage}{.24\textwidth}
        \centering
        \includegraphics[scale=0.12]{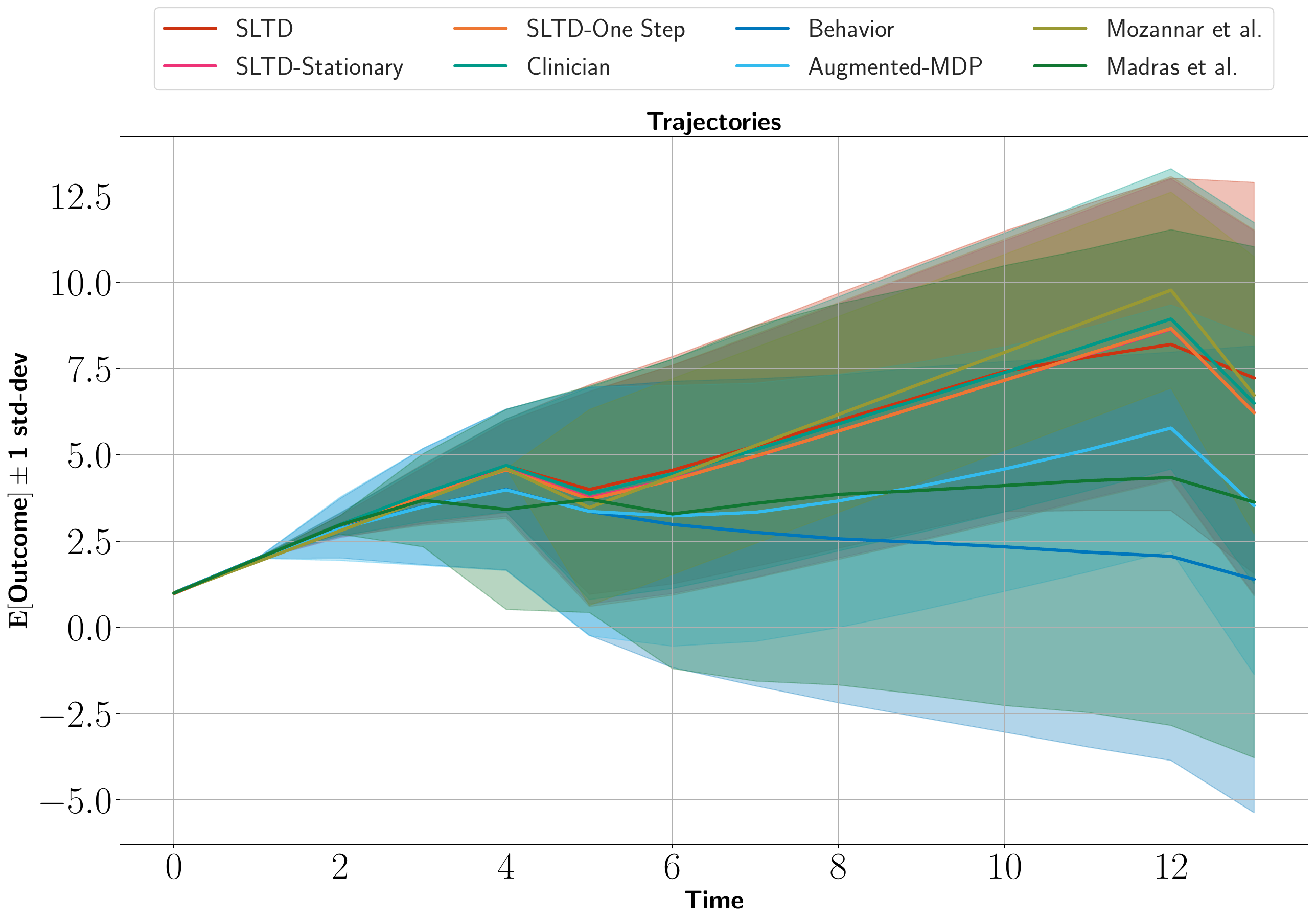}
        \caption*{{\small{\Synthetic}}}
        \label{fig:traj_disc}
    \end{minipage}%
    \begin{minipage}{0.24\textwidth}
        \centering
        \includegraphics[scale=0.12]{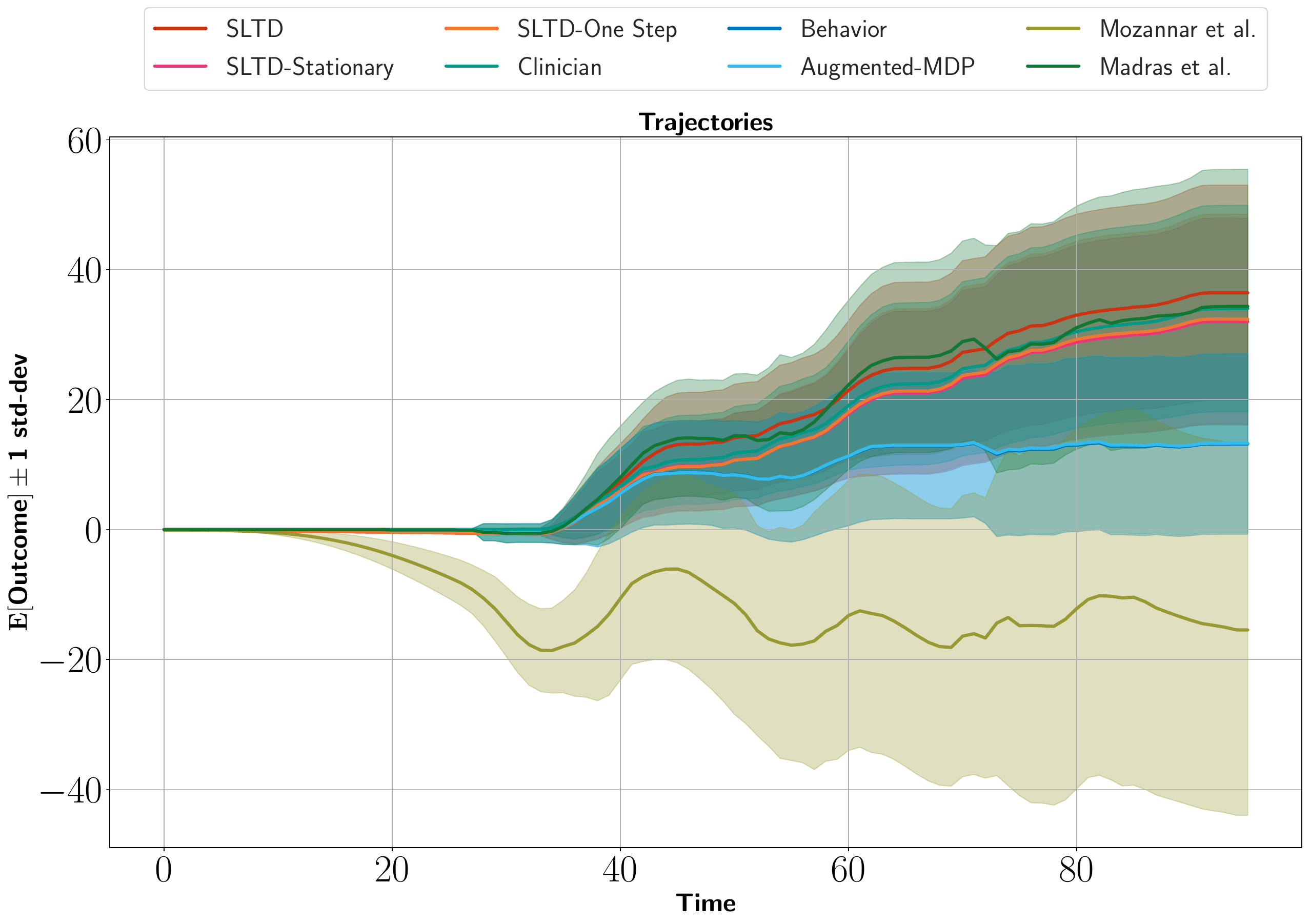}
       \caption*{{\small{Diabetes}}}
        \label{fig:traj_diab}
    \end{minipage}
       \begin{minipage}{0.24\textwidth}
        \centering
        \includegraphics[scale=0.12]{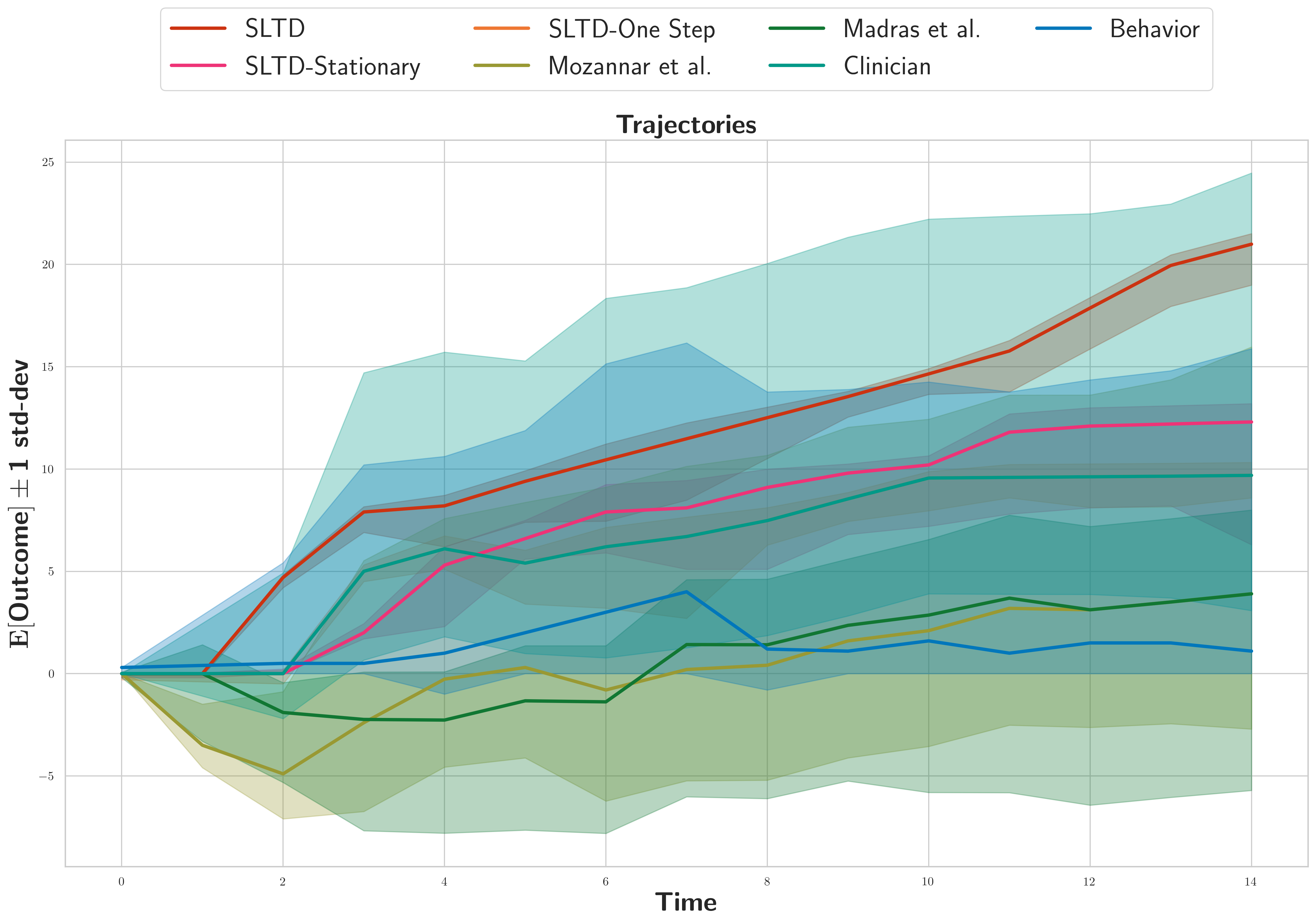}
       \caption*{{\small{HIV-I}}}
        \label{fig:traj_hiv_1}
    \end{minipage}
    \begin{minipage}{0.24\textwidth}
        \centering
        \includegraphics[scale=0.12]{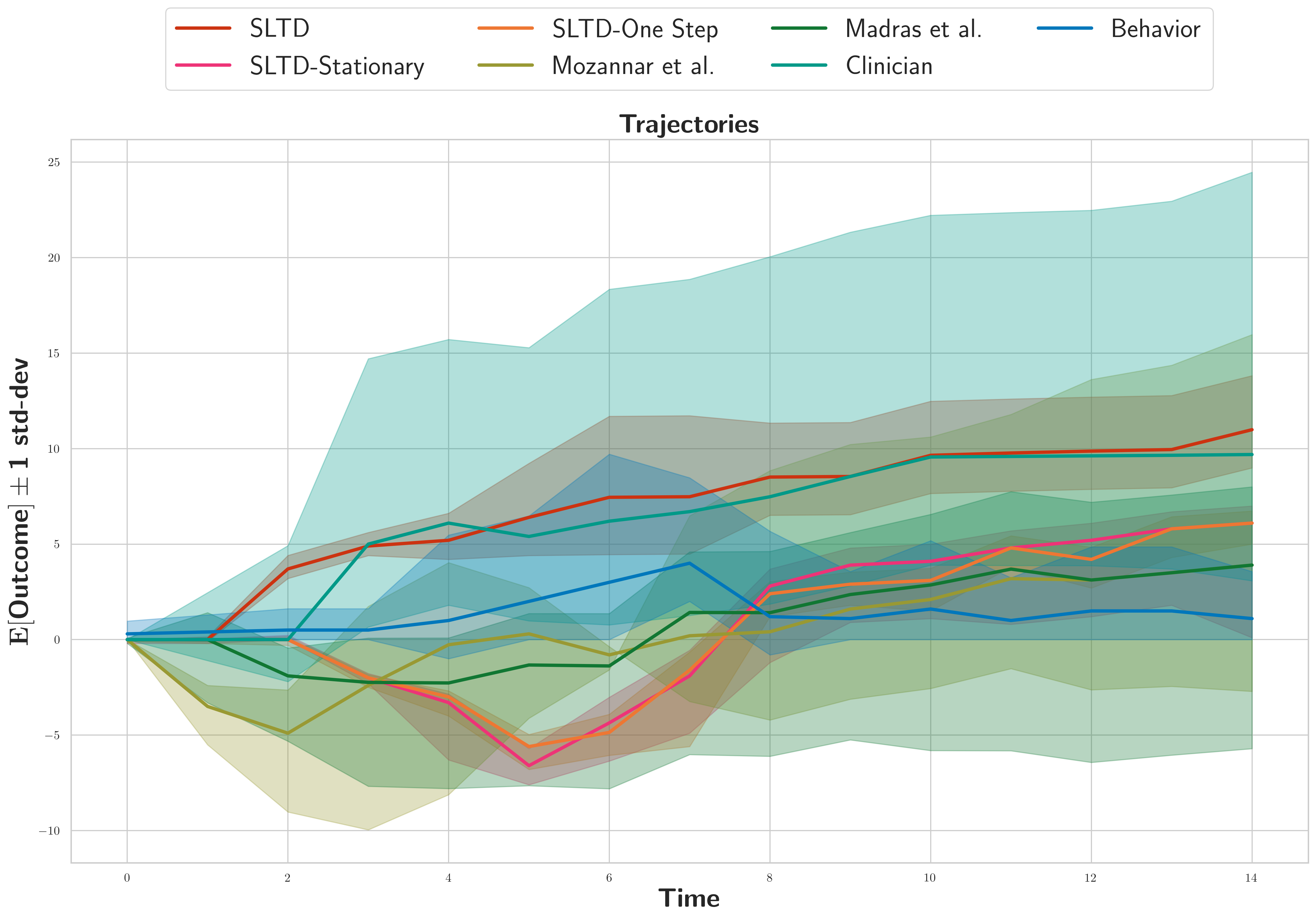}
       \caption*{{\small{HIV-II}}}
        \label{fig:traj_hiv_2}
    \end{minipage}
    \caption{Sample trajectories for \sltd variants, target policy $\pi_{\text{tar}}$ and clinician policy $\pi_{0}.$ Mean improvement are comparable for discrete data but significant using \sltd for Diabetes. Relative benefits of \sltd-stationary and \sltd-one-step are less compared to modeling non-stationarity particularly for  Diabetes data. Overall long-term uncertainty is comparable for all baselines for \Synthetic and Diabetes data.}
    \label{fig:trajectories}
\end{figure*}

\begin{figure*}[!htbp]
    \centering
    \begin{minipage}{\textwidth}
    \centering
    \includegraphics[scale=0.16]{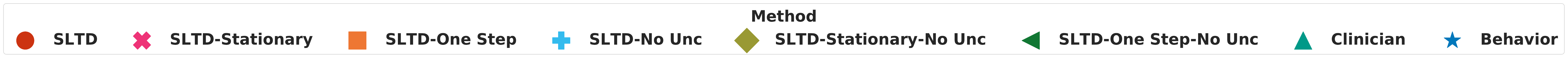}
    \end{minipage}
    \begin{minipage}{.24\textwidth}
        \centering
      \includegraphics[scale=0.10]{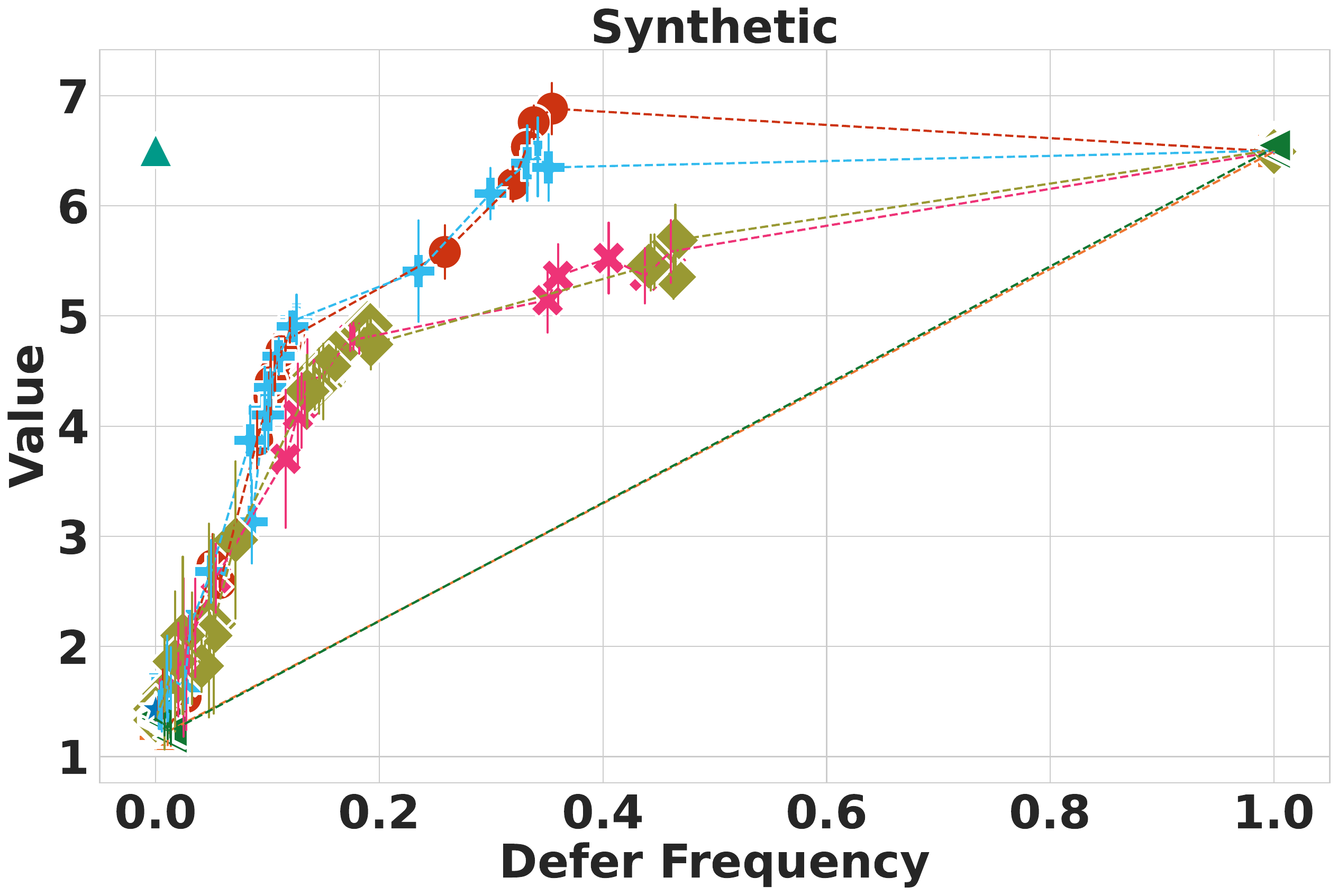}
        \label{fig:unc_decomp_toy}
    \end{minipage}%
    \begin{minipage}{0.24\textwidth}
        \centering
        \includegraphics[scale=0.10]{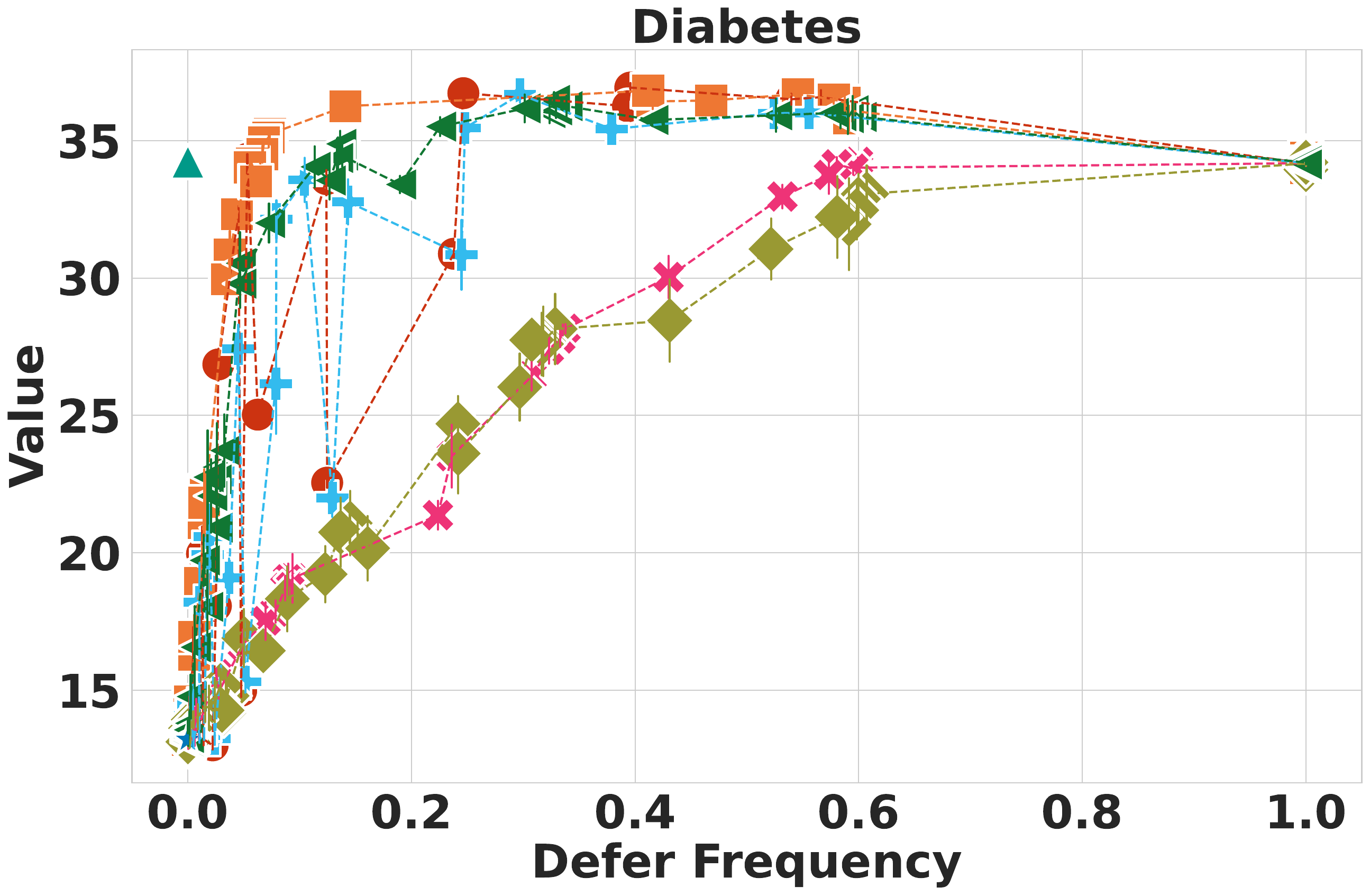} 
         \label{fig:unc_decomp_diabetes}
    \end{minipage}%
    \begin{minipage}{0.24\textwidth}
        \centering
        \includegraphics[scale=0.17]{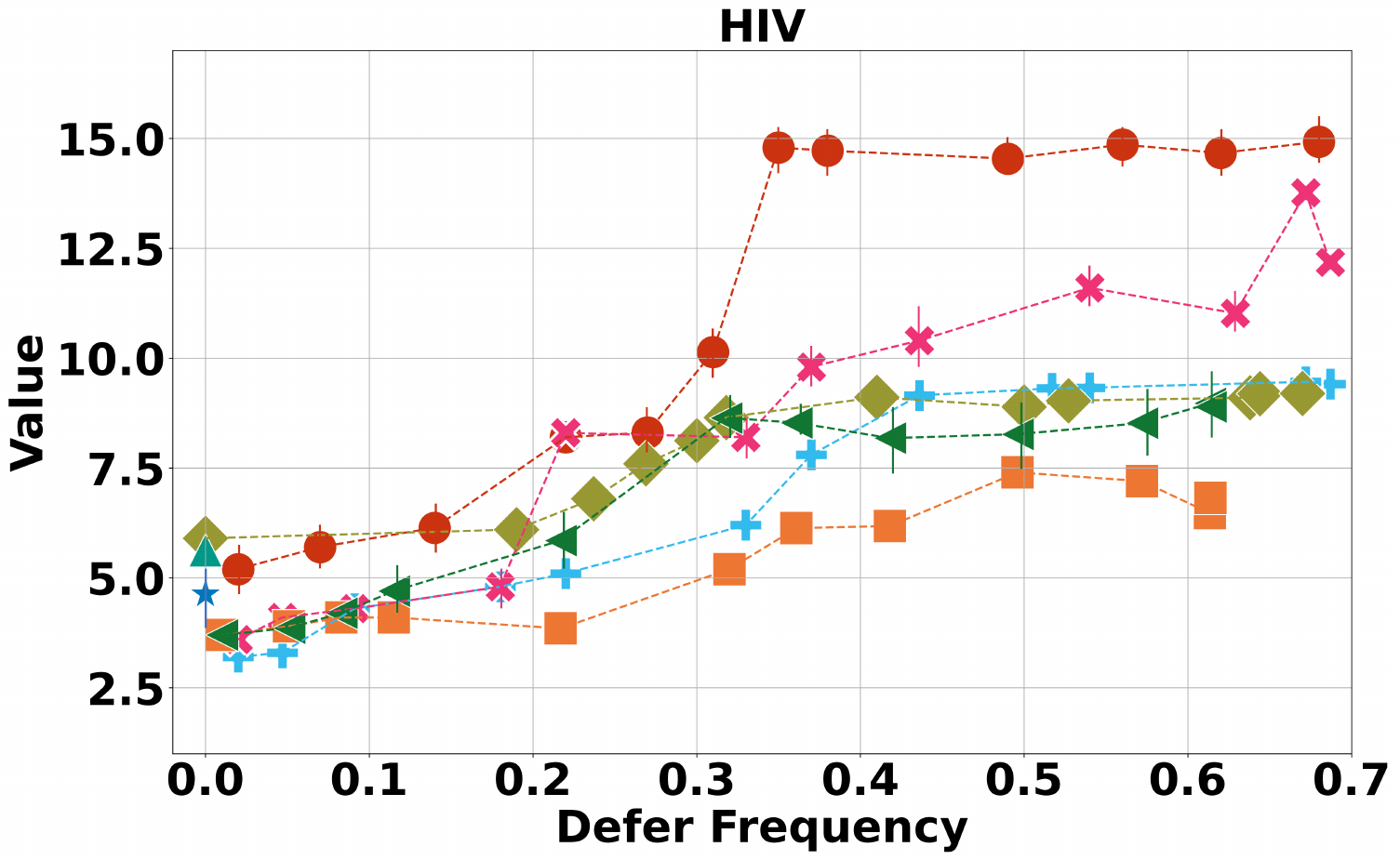} 
         \label{fig:unc_decomp_hiv1}
    \end{minipage}%
    \begin{minipage}{0.24\textwidth}
        \centering
        \includegraphics[scale=0.185]{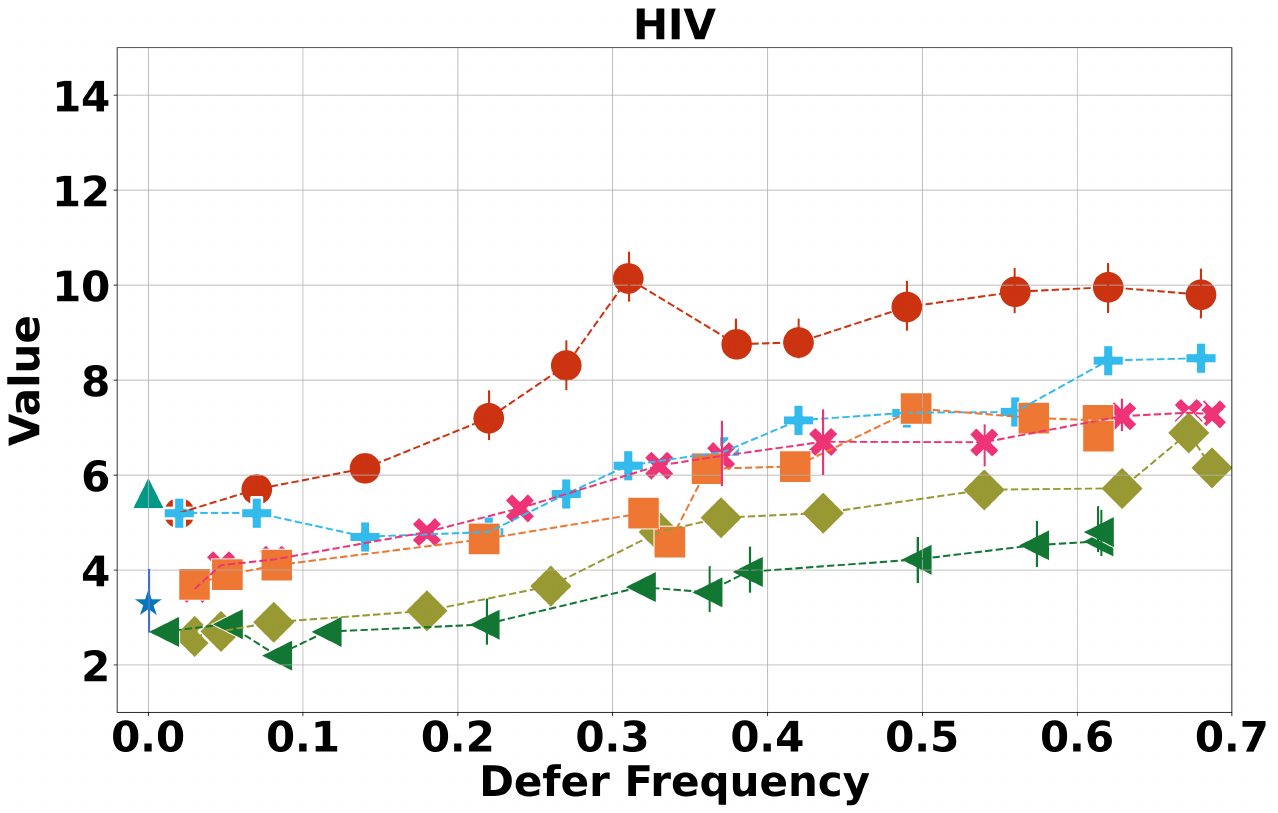} 
         \label{fig:unc_decomp_hiv2}
    \end{minipage}
    \caption{Trade-off of deferral frequence and expected value for \sltd, its variants, compared to ablations that do not estimate modeling uncertainty (No Unc, corresponding to $K=1$ in Equation~\ref{eq:G_est}). All ablations closely follow their counterparts that leverage modeling uncertainty for \Synthetic and Diabetes. This suggests low modeling uncertainty in our environments. A higher discrepancy in performance as in HIV-Case I and Case II is indicative of higher modeling uncertainty which can occur in low data regimes. Thus accounting for this uncertainty is crucial for learning a reliable deferral policy. When we account for this uncertainty, as in the original \sltd formulations, performance is significantly better for real-world HIV data.}
    \label{fig:uncertainty_ablations}
\end{figure*}

\paragraph{Evaluating learned deferral policy.}
To qualitatively analyze our policies, we plot the stochastic policies learned using \sltd, \sltd-One Step, \sltd Stationary in the following. Note that our policy function is $g_{\pi_{\text{tar}}}(s,t)$ is non-stationary. To visualize, we plot the probability of deferral over time. Figures~\ref{fig:diabetes_policies}, and \ref{fig:HIV_policies}, \ref{fig:HIV2_policies} shows the learned deferral policy for \sltd, \sltd-Stationary, and \sltd-One Step for Diabetes and HIV data respectively. As can be seen in Figure~\ref{fig:diabetes_policies}, as the transition dynamics shift over time to that of an alternative patient, the probability of deferral increases. While \sltd and \sltd-One Step show similar learned policies qualitatively, \sltd-Stationary variant defers in different states for Diabetes data giving a sense of issues due to misspecification of the dynamics. Note that \sltd-One step does not misspecify the dynamics but defers myopically. Similarly, for HIV Case-I and II in Figures~\ref{fig:HIV_policies} and \ref{fig:HIV2_policies}, the transition dynamics change over time as the virus evolves. Based on this evolution, the probability of deferral increases. In Case II, this evolution is more rapid, thus increasing the probability of deferral earlier than for Case I. Moreover, as the virus continues to mutate, the probability of deferral at subsequent steps increases for a sustained period of time in comparison to Case-I. This is typical of patients with many recombinant forms of the virus that evolve rapidly and have to be treated with more complex antiretroviral combinations.

Finally, we also notice certain differences across the two case studies for HIV. In the first case, it is evident that deferring to second line treatment in response to resistance from a first line treatment is helpful based on the results in Table \ref{tab:value_results} of the main paper. We further investigated whether deferral and points of high uncertainty correspond to certain events. In general, we observed that a higher probability of deferral and increased uncertainty at points of either virologic failure or where drug resistance has occurred. This is plausible as a change in therapy is typically required at this point to overcome such resistance. Unlike in Case I, Case II is significantly more challenging as it focuses on different viral strains that typically have a higher rate of evolution. Here, a higher probability of deferral is sustained across a longer time frame. Moreover, non-stationarity plays a significant role in this case, and the performance difference between SLTD and methods that do not account for this non-stationarity is more apparent.

{\bf{Sample Trajectories.}} Figure~\ref{fig:trajectories} shows sample trajectories (with uncertainty)  for all baselines and datasets to provide a sense of how the different baselines fare along with the long-term uncertainties. While no major differences between propagated uncertainties of \sltd variants is observed, the uncertainty can be higher for \mozannar especially for Diabetes data. This also applies for both cases of HIV.

{\bf{Additional Analysis of Uncertainty Ablations.}} Table~\ref{tab:value_results} shows the summary results corresponding to best performing parameters for all baselines. In addition, we also include uncertainty baselines corresponding to $K=1$. While Table~\ref{tab:value_results} only shows the best performing across all costs and other parameters, Figure~\ref{fig:uncertainty_ablations} shows the frequency-value trade-off for all parameter settings. For \Synthetic and Diabetes data, ablations suggest that there is no significant modeling uncertainty in our framework as the $K=1$ or ``No Unc'' counterparts closely follow the performance of $K=200$. This mainly suggests our modeling assumptions are reasonable and there is sufficient data to estimate the parameters of the dynamics resulting in low modeling uncertainty, and less variability across choice of $K$. This analysis can be done by collecting value estimates on true dynamics of the data for Discrete Toy and Diabetes. For HIV data, we obtain value estimates on maximum-likelihood estimates of the dynamics since true dynamics are unavailable for real-world data. Thus the analysis may be biased if the ML-estimate is biased. Nonetheless the significant difference suggests that there is indeed modeling uncertainty in the system for this data. Accounting for this uncertainty can thus have a significant impact on the  long-term outcomes as it will result in potentially delayed deferrals.

\end{document}